\documentclass[twoside,11pt]{article}
\usepackage{pgm}

\usepackage[utf8]{inputenc}
\inputencoding{utf8}

\usepackage{amssymb}

\usepackage{amsmath}
\usepackage{bm}
\usepackage{booktabs}
\usepackage{hyperref}
\usepackage{natbib}
\usepackage{lineno}
\usepackage{verbatim}
\usepackage{pgf, tikz}
\usetikzlibrary{arrows,automata,fit}
\usepackage{subfig}
\usepgflibrary{shapes.geometric}

\newcommand\independent{\protect\mathpalette{\protect\independenT}{\perp}}
\def\independenT#1#2{\mathrel{\rlap{$#1#2$}\mkern2mu{#1#2}}}

\newcommand{\Pa}{\operatorname{Pa}}

\newcommand{\xx}{1}
\newcommand{\yy}{1}

\newcommand{\stage}[2]{\tikz{\node[shape=circle,draw,inner sep=1pt,fill=#1]{$v_{#2}$};}}

\newcommand{\stages}[2]{\tikz{\node[shape=circle,draw,inner sep=1pt,fill=#1,minimum size=0.5cm]{\scriptsize{$v_{#2}$}};}} 

\newcommand{\leaf}{\tikz{\node[shape=circle,draw,inner sep=1.5pt,fill=white] {};}}

\ShortHeadings{A new class of generative classifiers based on staged tree models}{Carli, Leonelli and Varando}

\begin{document}

\title{A new class of generative classifiers based on staged tree models}
\author{\Name{Federico Carli} \Email{carli@dima.unige.it}\\
\addr Dipartimento di Matematica, Universit\`{a} degli Studi di Genova, Genova, Italy \and
\Name{Manuele Leonelli} \Email{manuele.leonelli@ie.edu}\\
   \addr School of Science and Technology, IE University, Madrid, Spain \and
   \Name{Gherardo Varando} \Email{gherardo.varando@uv.es}\\
   \addr Image Processing Laboratory, Universitat de Val\`{e}ncia, Val\`{e}ncia, Spain}

\maketitle

\begin{abstract}
Generative models for classification use the joint probability distribution of the class variable and the features to construct a decision rule. Among generative models, Bayesian networks and naive Bayes classifiers are the most commonly used and provide a clear graphical representation of the relationship among all variables. However, these have the disadvantage of highly restricting the type of relationships that could exist, by not allowing for context-specific independences. Here we introduce a new class of generative classifiers, called staged tree classifiers, which formally account for context-specific independence. They are constructed by a partitioning of the vertices of an event tree from which conditional independence can be formally read. The naive staged tree classifier is also defined, which extends the classic naive Bayes classifier whilst retaining the same complexity. An extensive simulation study shows that the classification accuracy of staged tree classifiers is competitive with that of state-of-the-art classifiers and an example showcases their use in practice.
\end{abstract}

\section{Introduction}
\label{sec:introduction}
The aim of statistical classification is to assign labels to instances described by  a vector of feature variables. The classification task is guided by a statistical model learnt using data containing labelled instances. The array of models designed to perform classification is constantly increasing and includes, among others, random forests \citep{Ho1995}, recursive partitioning \citep{Breiman1984} and probabilistic neural networks \citep{Specht1990}.

Bayesian network classifiers (BNCs) \citep{Bielza2014,Friedman1997} are special types of Bayesian networks (BNs) designed for classification problems. These have been applied to a wide array of real-world applications with competitive classification performance against state-of-the-art classifiers \citep{Flores2012}. There are numerous advantages associated to BNCs. First, they provide an  explicit and intuitive representation of the relationship among features represented by a graph. Second, they are a fully coherent probabilistic model thus giving uncertainty measures about the chosen labels. Third, many of the methods and algorithms developed for general BNs can be simply adapted and used for BNCs \citep[see e.g.][]{Benjumeda2019}. Last, they are implemented in various pieces of user-friendly software \citep[see e.g. the \texttt{bnclassify} \texttt{R} package of][]{Mihaljevic2018}. More generally, BNCs are generative classifiers which give an estimate of the joint distribution of both the features and the class.

One of the main limitations of BNs is that they can only explicitly represent symmetric conditional independences among variables of interest. However, in many applied domains, conditional independences are context-specific, meaning that they only hold for specific instantiations of the conditioning variables. For this reason, numerous extensions of BNs have been proposed that can take into account asymmetric independences \citep{Boutilier1996,Cano2012,Jaeger2006,Pensar2015,Pensar2016,Poole2003}. With the exception of \citet{Jaeger2006} and \citet{Pensar2015}, all these models somehow lose the intuitiveness of BNs since they cannot represent all the models' information into a unique graph.

Despite the efforts in accommodating asymmetries in BNs, the development of BNCs embedding context-specific information has been limited. Solutions proposed  to address this issue use the idea of Bayesian multinets \citep{Geiger1996}, which consist of several networks each associated with a subset of the domain of one variable, often called distinguished. Bayesian multinets have been used for classification in \citet{Friedman1997}, \citet{Gurwicz2006}, \citet{Huang2003} and \citet{Hussein2004}, among others.

In this paper a novel class of generative classifiers based on staged trees \citep{Collazo2018,Smith2008} embedding asymmetric conditional independences is considered, and henceforth called \emph{staged tree classifiers}. Staged trees are defined as probability trees \citep{Shafer1996} embellished with asymmetric conditional independence information, represented by a partitioning of the vertices of the tree. Staged tree classifiers are staged trees whose topology is specifically designed for classification problems. Although staged trees have been used in a variety of applications, including the modelling of health problems \citep{Barclay2013, Keeble2017} and criminal activities \citep{Collazo2016}, their specific use for classification problems has been limited. 

Staged tree classifiers are generative classifiers which, whilst extending the class of BNCs to deal with asymmetric conditional independences, share the same advantages of  BNCs: first, the relationship between the random variables is still intuitively depicted in a unique graph; second, they are fully coherent probabilistic models; third, learning algorithms already defined for staged trees \citep[e.g.][]{Freeman2011,Silander2013,Leonelli2022a,Leonelli2022b} can simply be adapted for classification purposes; fourth, the freely-available \texttt{R} package \texttt{stagedtrees} \citep{Carli2020} gives an implementation of a variety of learning algorithms and inferential routines to apply the methods in practice. 


Mirroring the theory of BNCs, staged tree classifiers of different complexity are discussed and their properties investigated. Notably, the naive staged tree classifier is introduced which is shown to have the same complexity of the standard naive Bayes classifier, but relaxing the strict conditional independence assumptions of naive Bayes. Experimental studies demonstrate that staged tree classifiers have comparable classification rates to state-of-the-art classifiers and outperform other generative classifiers.

The paper is structured as follows. Section \ref{sec:bnc} introduces the notation and reviews BNCs. Staged trees are reviewed in Section \ref{sec:ceg}. Staged tree classifiers are introduced and studied in Section \ref{sec:cegc}. Section \ref{sec:learning} discusses learning algorithms. Section \ref{sec:experimental} presents an experimental study and Section \ref{sec:applied} discusses a classification application in details. The paper is concluded with a discussion.

\section{Bayesian network classifiers}
\label{sec:bnc}

\subsection{The Bayesian network model}

Let $G=([p],E_G)$ be a directed acyclic graph (DAG) with vertex set $[p]=\{1,\dots,p\}$ and edge set $E_G$. Let $\bm{X}=(X_i)_{i\in[p]}$ be categorical random variables with joint mass function $P$ and sample space $\mathbb{X}=\times_{i\in[p]}\mathbb{X}_i$. For $A\subset [p]$, we let $\bm{X}_A=(X_i)_{i\in A}$ and $\bm{x}_A=(x_i)_{i\in A}$ where $\bm{x}_A\in\mathbb{X}_A=\times_{i\in A}\mathbb{X}_i$. We say that $P$ is Markov to $G$ if, for $\bm{x}\in\mathbb{X}$, 
\begin{equation}
\label{eq:markov}
P(\bm{x})=\prod_{k\in[p]}P(x_k | \bm{x}_{\Pi_k}),
\end{equation}
where $\Pi_k$ is the parent set of $k$ in $G$ and $P(x_k | \bm{x}_{\Pi_k})$ is a shorthand for $P(X_k=x_k |\bm{X}_{\Pi_k} = \bm{x}_{\Pi_k})$. Henceforth, we assume the existence
of a linear ordering $\sigma$ of $[p]$ for which only pairs $(i,j)$ where $i$ appears before $j$ in the order can be in the edge set.

The ordered Markov condition implies conditional independences of the form
\begin{equation}
\label{ci}
X_i \independent \bm{X}_{[i-1]}\,|\, \bm{X}_{\Pi_i}.
\end{equation}
\begin{definition}
Let $G$ be a DAG and $P$ Markov to $G$. The \emph{Bayesian network} model (associated to $G$) is 
\[
\mathcal{M}_G = \{P\in\Delta_{|\mathbb{X}|-1}\,|\, P \mbox{ is Markov to } G\}.
\]
where $\Delta_{|\mathbb{X}|-1}$ is the ($|\mathbb{X}|-1$)-dimensional  probability simplex.
\end{definition}

\begin{figure}
\begin{center}
\begin{tikzpicture}[
            > = stealth, 
            shorten > = 1pt, 
            auto,
            node distance = 2cm, 
            semithick 
        ]

        \tikzstyle{every state}=[
            draw = black,
            thick,
            fill = white,
            minimum size = 8mm
        ]
\renewcommand{\xx}{2}
\renewcommand{\yy}{1.5}
\node[state] (1) at (0*\xx,0*\yy){1};
\node[state] (2) at (0*\xx,1*\yy){2};
\node[state] (3) at (1*\xx,0*\yy){3};
\node[state] (4) at (1*\xx,1*\yy){4};
\node[state] (5) at (2*\xx,0.5*\yy){5};
\draw[->, line width = 1.1pt] (1) -- (3);
\draw[->, line width = 1.1pt] (2) -- (4);
\draw[->, line width = 1.1pt] (3) -- (5);
\draw[->, line width = 1.1pt] (4) -- (5);
\draw[->, line width = 1.1pt] (1) -- (4);
\end{tikzpicture}
\caption{A simple DAG with vertex set $\{1,2,3,4,5\}$ and edge set $\{(1,3),(1,4),(2,4),(3,5),(4,5)\}$.\label{fig:bn}}
\end{center}
\end{figure}
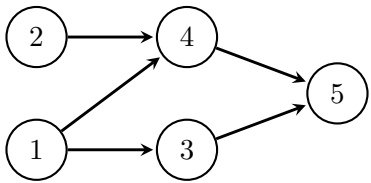

Figure \ref{fig:bn} reports a simple DAG with five vertices. Any Markov distribution to this DAG must factorize as
$P(x_5|x_3,x_4)P(x_4|x_1,x_2)P(x_3|x_1,x_2)P(x_2)P(x_1).$

Let $\mathcal{G}$ be the set of DAGs with vertex set $[p]$ and ordering $\sigma$. We define the space of BN models over $\bm{X}$ as $\mathcal{M}_{\mathcal{G}}=\cup_{G\in\mathcal{G}}\mathcal{M}_G$

\subsection{Bayesian networks for classification}
Supppose now that $\bm{X}=(X_1,\dots,X_p)$ is a $p$-dimensional vector of categorical feature variables and $C$ a categorical class variable with sample space $\mathbb{C}$ and $c\in\mathbb{C}$. Given a training set of labelled observations $\mathcal{D}=\{(\bm{x}^1,c^1),\dots,$ $(\bm{x}^N,c^N)\}$, where  $\bm{x}^i\in\mathbb{X}$ and $c^i\in\mathbb{C}$, the aim of a generative classifier is to learn a joint probability $p(c,\bm{x})$ and  assign a non-labelled instance $\bm{x}$ to the most probable a posteriori class found as
\[
\arg\max_{c\in\mathbb{C}}p(c|\bm{x})=\arg\max_{c\in\mathbb{C}}p(c,\bm{x}).
\]
Such a classifier is referred to as \textit{Bayes classifier}.

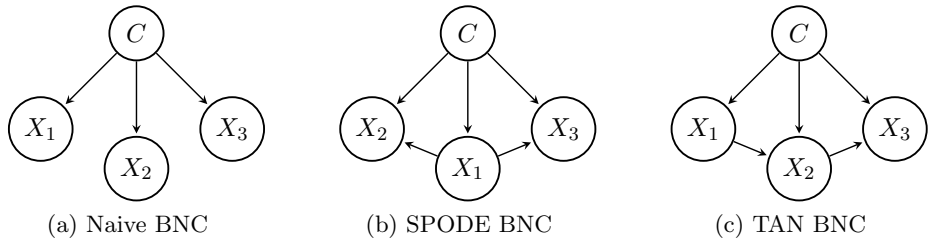
\begin{figure}
\centering

\subfloat[Naive BNC \label{fig:naivebn}]{
\scalebox{0.9}{
\begin{tikzpicture}[
            > = stealth, 
            shorten > = 1pt, 
            auto,
            node distance = 2cm, 
            semithick 
        ]

        \tikzstyle{every state}=[
            draw = black,
            thick,
            fill = white,
            minimum size = 8mm
        ]

        \node[state] (c) {$C$};
        \node[state] (x1) [below left of=c] {$X_1$};
        \node[state] (x2) [below of=c] {$X_2$};
        \node[state] (x3) [below right of=c] {$X_3$};
        
        \path[->] (c) edge   (x1);
        \path[->] (c) edge   (x2);
        \path[->] (c) edge   (x3);
  
 \end{tikzpicture}
}}
\hspace{0.5cm}
\subfloat[SPODE BNC \label{fig:spodebn}]{ 
\scalebox{0.9}{
 \begin{tikzpicture}[
            > = stealth, 
            shorten > = 1pt, 
            auto,
            node distance = 2cm, 
            semithick 
        ]

        \tikzstyle{every state}=[
            draw = black,
            thick,
            fill = white,
            minimum size = 8mm
        ]

        \node[state] (c) {$C$};
        \node[state] (x2) [below left of=c] {$X_2$};
        \node[state] (x1) [below of=c] {$X_1$};
        \node[state] (x3) [below right of=c] {$X_3$};
        
        \path[->] (c) edge   (x1);
        \path[->] (c) edge   (x2);
        \path[->] (c) edge   (x3);
         \path[->] (x1) edge   (x2);
        \path[->] (x1) edge   (x3);
        
 \end{tikzpicture}
}}
\hspace{0.5cm}
 \subfloat[TAN BNC \label{fig:tanbn}]{   
 \scalebox{0.9}{
 \begin{tikzpicture}[
            > = stealth, 
            shorten > = 1pt, 
            auto,
            node distance = 2cm, 
            semithick 
        ]

        \tikzstyle{every state}=[
            draw = black,
            thick,
            fill = white,
            minimum size = 8mm
        ]

        \node[state] (c) {$C$};
        \node[state] (x1) [below left of=c] {$X_1$};
        \node[state] (x2) [below of=c] {$X_2$};
        \node[state] (x3) [below right of=c] {$X_3$};
        
        \path[->] (c) edge   (x1);
        \path[->] (c) edge   (x2);
        \path[->] (c) edge   (x3);
        \path[->] (x1) edge   (x2);
        \path[->] (x2) edge   (x3);
 \end{tikzpicture}
}}
\caption{Examples of BNCs with three features and one class. \label{fig:bnc}}
\end{figure}

BNCs are Bayes classifiers that factorize $p(c,\bm{x})$ according to a BN over the variables $X_1,\dots,X_p$ and $C$. Although any BN model could be used for classification purposes, most often the underlying DAG is restricted so that the class variable $C$ has no parents. Therefore, in BNCs the class variable is the root of the DAG and there is a direct link from $C$ to $X_i$, for $i\in [p]$, since otherwise features not connected to the class would not provide any information for classification. The simplest possible model is the so-called naive Bayes classifier \citep{Minsky1961} which assumes the features are conditionally independent given the class (Figure \ref{fig:naivebn}). BNCs of increasing complexity can then be defined by adding dependences between the feature variables. For instance, the super-parent-one-dependence-estimator (SPODE) BNC \citep{Keogh2002} assumes there is a feature parent of all others (Figure \ref{fig:spodebn}). Another commonly used classifier is the tree-augmented naive (TAN) BNC \citep{Friedman1997} for which each feature has at most two parents: the class and possibly another feature (Figure \ref{fig:tanbn}).  

Although BNCs of any complexity can be learnt and used in practice, empirical evidence demonstrates that model complexity does not necessarily implies better classification accuracy \citep{Bielza2014}. Despite of their simplicity, naive BNCs have been shown to lead to good accuracy in classification problems \citep{Bielza2014,Flores2012}.

Alongside these empirical evaluations of BNCs, theoretical studies about the 
expressiveness of such models have appeared. Recently, \citet{Varando2015} and 
\citet{Varando2016} fully characterized the decision functions induced by 
various BNCs and consequently derived bounds for their expressive power. 
They built on the work of \citet{Ling2002} that demonstrated that any BNC whose 
vertices have at most $k$ parents cannot represent any decision function 
containing $(k+1)$-XORs \citep[also known as parity functions][]{Donnell2014}. 
Thus naive BNCs are not capable of capturing any 2-XORs. 
On the positive side, \citet{Domingos1997} demonstrated that naive BNCs are optimal 
under a $0-1$ loss even when the assumption of conditional independence 
among features does not hold.

\section{Staged trees}
\label{sec:ceg}

Because of the strict assumptions on the symmetry of independences among variables in
BNs, models accommodating asymmetric conditional independence statements have been 
developed. 
Staged trees are one extension of BNs whose conditional independences can 
still directly be read from the associated graphical 
representation \citep{Collazo2018,Smith2008}. 
However, differently to BNs, whose graphical representation is a DAG, 
staged trees are constructed from probability trees as detailed next.

\subsection{$X$-compatible staged trees}

Consider a $p$-dimensional random vector $\bm{X}$ taking values in the product sample space $\mathbb{X}$. Let $(V,E)$ be a directed, finite, rooted tree with vertex set $V$, root node $v_0$ and edge set $E$. 
For each $v\in V$, 
let $E(v)=\{(v,w)\in E\}$ be the set of edges emanating
from $v$ and $\mathcal{L}$ be a set of labels. 

\begin{definition}
\label{def:x}
An $\bf X$-compatible staged tree 
is a triple $(V,E,\theta)$, where $(V,E)$ is a rooted directed tree and:
\begin{enumerate}
    \item $V = {v_0} \cup \bigcup_{i \in [p]} \mathbb{X}_{[i]}$;
		\item For all $v,w\in V$,
$(v,w)\in E$ if and only if $w=\bm{x}_{[i]}\in\mathbb{X}_{[i]}$ and 
			$v = \bm{x}_{[i-1]}$, or $v=v_0$ and $w=x_1$ for some
$x_1\in\mathbb{X}_1$;
\item $\theta:E\rightarrow \mathcal{L}^*=\mathcal{L}\times \cup_{i\in[p]}\mathbb{X}_i$ is a labelling of the edges such that $\theta(v,\bm{x}_{[i]}) = (\kappa(v), x_i)$ for some 
			function $\kappa: V \to \mathcal{L}$. The function $k$ is called the colouring of the staged tree $T$.
\end{enumerate}
	If $\theta(E(v)) = \theta(E(w))$ then $v$ and $w$ are said to be in the same 	\emph{stage}.
\end{definition} 

Therefore, the equivalence classes induced by  $\theta(E(v))$
form a partition of the internal vertices of the tree  in \emph{stages}.

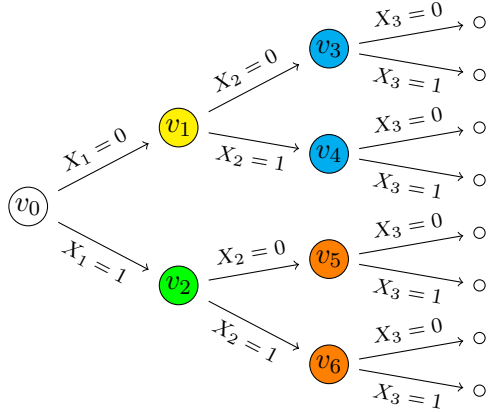
\begin{figure}
\centering
\begin{tikzpicture}
\renewcommand{\xx}{2}
\renewcommand{\yy}{0.7}
\node (v1) at (0*\xx,0*\yy) {\stage{white}{0}};
\node (v2) at (1*\xx,1.5*\yy) {\stage{yellow}{1}};
\node (v3) at (1*\xx,-1.5*\yy) {\stage{green}{2}};
\node (v4) at (2*\xx,3*\yy) {\stage{cyan}{3}};
\node (v5) at (2*\xx,1*\yy) {\stage{cyan}{4}};
\node (v6) at (2*\xx,-1*\yy) {\stage{orange}{5}};
\node (v7) at (2*\xx,-3*\yy) {\stage{orange}{6}};
\node (l1) at (3*\xx,3.5*\yy) {\leaf};
\node (l2) at (3*\xx,2.5*\yy) {\leaf};
\node (l3) at (3*\xx,1.5*\yy) {\leaf};
\node (l4) at (3*\xx,0.5*\yy) {\leaf};
\node (l5) at (3*\xx,-0.5*\yy) {\leaf};
\node (l6) at (3*\xx,-1.5*\yy) {\leaf};
\node (l7) at (3*\xx,-2.5*\yy) {\leaf};
\node (l8) at (3*\xx,-3.5*\yy) {\leaf};
\draw[->] (v1) -- node [above, sloped] {\scriptsize{$X_1=0$}} (v2);
\draw[->] (v1) -- node [below, sloped] {\scriptsize{$X_1=1$}} (v3);
\draw[->] (v2) --  node [above, sloped] {\scriptsize{$X_2=0$}} (v4);
\draw[->] (v2) -- node [below, sloped] {\scriptsize{$X_2=1$}} (v5);
\draw[->] (v3) -- node [above, sloped] {\scriptsize{$X_2=0$}} (v6);
\draw[->] (v3) -- node [below, sloped] {\scriptsize{$X_2=1$}} (v7);
\draw[->] (v4) -- node [above, sloped] {\scriptsize{$X_3=0$}} (l1);
\draw[->] (v4) -- node [below, sloped] {\scriptsize{$X_3=1$}} (l2);
\draw[->] (v5) -- node [above, sloped] {\scriptsize{$X_3=0$}} (l3);
\draw[->] (v5) -- node [below, sloped] {\scriptsize{$X_3=1$}} (l4);
\draw[->] (v6) -- node [above, sloped] {\scriptsize{$X_3=0$}} (l5);
\draw[->] (v6) -- node [below, sloped] {\scriptsize{$X_3=1$}} (l6);
\draw[->] (v7) -- node [above, sloped] {\scriptsize{$X_3=0$}} (l7);
\draw[->] (v7) -- node [below, sloped] {\scriptsize{$X_3=1$}} (l8);
\end{tikzpicture}

\caption{An example of an $\bf X$-compatible staged tree. \label{fig:staged1}}
\end{figure}

Definition \ref{def:x} first constructs a rooted tree where each root-to-leaf path, or equivalently each leaf, is associated to an element of the sample space $\mathbb{X}$.  Then a labeling of the edges of such a tree is defined where labels are pairs with one element from a set $\mathcal{L}$ and the other from the sample space $\mathbb{X}_i$ of the corresponding variable $X_i$ in the tree. By construction, $\bf X$-compatible staged trees are such that two vertices can be in the same stage if and only if they correspond to the same sample space. Although staged trees can be more generally defined without imposing this condition \citep[see e.g.][]{Collazo2018}, henceforth, and as common in practice, we focus on $\bf{X}$-compatible staged trees only \citep[see][for an example of a non $\bf{X}$-compatible tree]{Leonelli2019}. 

Figure \ref{fig:staged1} reports an $(X_1,X_2,X_3)$-compatible staged tree over three binary variables. The \textit{coloring} given by the function $\kappa$ is shown in the vertices and
each edge $(\cdot , (x_1, \ldots, x_{i}))$ is labeled with $X_i = x_{i}$. 
The edge labeling $\theta$ can be read from the graph combining the text label and the 
color of the emanating vertex. 
The staging of the staged tree in Figure \ref{fig:staged1} is given by the partition $\{v_0\}$, $\{v_1\}$, $\{v_2\}$, $\{v_3,v_4\}$ and $\{v_5,v_6\}$.

The parameter space associated to an $\bf X$-compatible staged tree $T = (V, E, \theta)$ 
with 
labeling $\theta:E\rightarrow \mathcal{L}^{*}$ 
is defined as
\begin{equation}
\label{eq:parameter}
	\Theta_T=\Big\{\bm{y}\in\mathbb{R}^{|\theta(E)|} \;|\; \forall ~ e\in E, y_{\theta(e)}\in (0,1)\textnormal{ and }\sum_{e\in E(v)}y_{\theta(e)}=1\Big\}
\end{equation}
Equation~(\ref{eq:parameter}) defines a class of probability mass functions 
over the edges emanating from any internal vertex coinciding with conditional distributions  $P(x_i | \bm{x}_{[i-1]})$, $\bm{x}\in\mathbb{X}$ and $i\in[p]$. In the staged tree in Figure \ref{fig:staged1} the staging $\{v_3,v_4\}$ implies that the conditional distribution of $X_3$ given $X_1=0$, and $X_2 = 0$, represented by the edges emanating from $v_3$, is equal to the conditional distribution of $X_3$ given $X_1=0$ and $X_2=1$. A similar interpretation holds for the staging $\{v_5,v_6\}$. This in turn implies that  $X_3\independent X_2|X_1$, thus illustrating that the staging of a tree is associated to conditional independence statements.

Let $\bm{l}_{T}$ denote the leaves of a staged tree $T$. Given a vertex $v\in V$, there is a unique path in $T$ from the root $v_0$ to $v$, denoted as $\lambda(v)$. The number of edges in $\lambda(v)$ is called  the distance of $v$, and the set of vertices at distance $k$ is denoted by $V_k$.  For any path $\lambda$ in $T$, let $E(\lambda)=\{e\in E: e\in \lambda\}$ denote the set of edges in the path $\lambda$.

\begin{definition}
\label{def:stmodel}
	The \emph{staged tree model} $\mathcal{M}_{T,\theta}$ associated to the $\bf X$-compatible staged 
	tree $(V,E,\theta)$ is the image of the map
\begin{equation}
\label{eq:model}
\begin{aligned}
\phi_T & : \Theta_T\rightarrow \Delta_{|\bm{l}_T|-1}^{\circ};\\
\phi_T & : y \mapsto p_l = \Big(\prod_{e\in E(\lambda(l))}y_{\theta(e)}\Big)_{l\in \bm{l}_T}
\end{aligned}
\end{equation}
\end{definition}
Therefore, staged trees models are such that atomic probabilities are equal to the product of the edge labels in root-to-leaf paths and coincide with the usual factorization of mass functions via recursive conditioning.

Let $\Theta$ be the set of functions $\theta$ from $E$ to $\mathcal{L}^{*}$, that is all possible partitions, or staging, of the staged tree. We define $\mathcal{M}_T=\cup_{\theta\in\Theta}\mathcal{M}_{T,\theta}$. So as $\mathcal{M}_{\mathcal{G}}$ is the union of all possible BN models given a specific ordering, $\mathcal{M}_T$ is the union of all possible staged tree models, that is of all possible stagings, given a specific ordering of the variables.

\subsection{Staged trees and Bayesian networks}
\label{sec:noia}

\citet{Smith2008} demonstrated that any BN can be represented as an equivalent staged tree, whilst the converse is not true. We next illustrate how to construct a staged tree equivalent to a BN.  Consider a DAG
$G$ and an $\bf X$-compatible staged tree with vertex set $V$,  
edge set $E$ and labeling $\theta$ defined via the 
coloring $\kappa(\bm{x}_{[i]} ) = \bm{x}_{\Pi_{i}}$ of the vertices. The staged tree $T_G$, with vertex set $V$, edge set $E$ and labeling $\theta$
so constructed, is called \emph{the staged tree model of $G$}. 
Importantly,
$\mathcal{M}_G= \mathcal{M}_{T_G,\theta}$, i.e. the two models are exactly the same,
since they entail exactly the same factorization of the joint
probability \citep{Varando2021}. The staging of $T_G$  represents the
Markov conditions associated to the graph $G$.

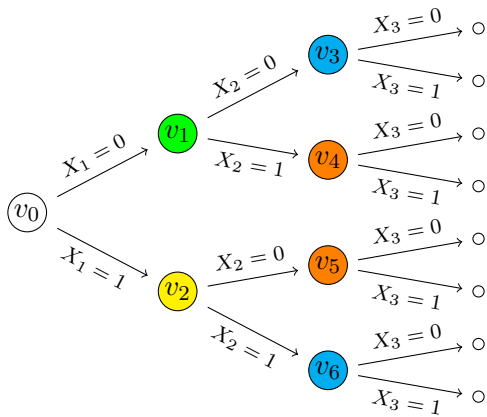
\begin{figure}
\centering
\begin{tikzpicture}
\renewcommand{\xx}{2}
\renewcommand{\yy}{0.7}
\node (v1) at (0*\xx,0*\yy) {\stage{white}{0}};
\node (v2) at (1*\xx,1.5*\yy) {\stage{green}{1}};
\node (v3) at (1*\xx,-1.5*\yy) {\stage{yellow}{2}};
\node (v4) at (2*\xx,3*\yy) {\stage{cyan}{3}};
\node (v5) at (2*\xx,1*\yy) {\stage{orange}{4}};
\node (v6) at (2*\xx,-1*\yy) {\stage{orange}{5}};
\node (v7) at (2*\xx,-3*\yy) {\stage{cyan}{6}};
\node (l1) at (3*\xx,3.5*\yy) {\leaf};
\node (l2) at (3*\xx,2.5*\yy) {\leaf};
\node (l3) at (3*\xx,1.5*\yy) {\leaf};
\node (l4) at (3*\xx,0.5*\yy) {\leaf};
\node (l5) at (3*\xx,-0.5*\yy) {\leaf};
\node (l6) at (3*\xx,-1.5*\yy) {\leaf};
\node (l7) at (3*\xx,-2.5*\yy) {\leaf};
\node (l8) at (3*\xx,-3.5*\yy) {\leaf};
\draw[->] (v1) -- node [above, sloped] {\scriptsize{$X_1=0$}} (v2);
\draw[->] (v1) -- node [below, sloped] {\scriptsize{$X_1=1$} } (v3);
\draw[->] (v2) --  node [above, sloped] {\scriptsize{$X_2=0$}} (v4);
\draw[->] (v2) -- node [below, sloped] {\scriptsize{$X_2=1$}} (v5);
\draw[->] (v3) -- node [above, sloped] {\scriptsize{$X_2=0$}} (v6);
\draw[->] (v3) -- node [below, sloped] {\scriptsize{$X_2=1$}} (v7);
\draw[->] (v4) -- node [above, sloped] {\scriptsize{$X_3=0$}} (l1);
\draw[->] (v4) -- node [below, sloped] {\scriptsize{$X_3=1$}} (l2);
\draw[->] (v5) -- node [above, sloped] {\scriptsize{$X_3=0$}} (l3);
\draw[->] (v5) -- node [below, sloped] {\scriptsize{$X_3=1$}} (l4);
\draw[->] (v6) -- node [above, sloped] {\scriptsize{$X_3=0$}} (l5);
\draw[->] (v6) -- node [below, sloped] {\scriptsize{$X_3=1$}} (l6);
\draw[->] (v7) -- node [above, sloped] {\scriptsize{$X_3=0$}} (l7);
\draw[->] (v7) -- node [below, sloped] {\scriptsize{$X_3=1$}} (l8);
\end{tikzpicture}

\caption{An example of a staged tree not associated to any BN $G$. \label{fig:staged2}}
\end{figure}

For instance, the staged tree in Figure \ref{fig:staged1} can be constructed as the $T_G$ from the BN with DAG $X_2\leftarrow X_1\rightarrow X_3$. Conversely, consider the staged tree in Figure \ref{fig:staged2}. The blue staging implies that the conditional distribution of $X_3$ given $X_2=X_1=0$ is equal to the conditional distribution of $X_3$ given $X_2=X_1=1$. Such a constraint cannot be explicitly represented by the DAG of a BN and therefore there is no DAG $G$ such that $\mathcal{M}_G=\mathcal{M}_{T_G,\theta}$, i.e. there is no BN which is equivalent to the staged tree in Figure \ref{fig:staged2}. More generally, it holds that $\mathcal{M}_{\mathcal{G}}\subset \mathcal{M}_T$ \citep{Varando2021}.

\section{Staged tree classifiers}
\label{sec:cegc}

The technology of staged trees has been refined over the years and methods to investigate causal  relationships \citep{deep,Thwaites2010}, perform statistical inference \citep{Gorgen2015},  check model's robustness \citep{Leonelli2017} and carry out causal discovery \citep{Leonelli2021} are now available. However, the specific use of staged trees for classification has not been investigated in the literature. Thus, just as BNCs have been defined as a specific subclass of BNs whose graph entertains some properties,  the class of staged tree classifiers is defined here. 

As in Section \ref{sec:bnc}, suppose $\bm{X}=(X_1,\dots,X_n)$ is a vector of features and $C$ is the class variable. 

\begin{definition}
A \textbf{staged tree classifier} for the class $C$ and features $\bm{X}$ 
	is a $(C, \bm{X})$-compatible  staged tree.
\end{definition}

The requirement of $C$ being the root of the tree follows from the idea that in most BNCs the class has no parents, so to maximise the information provided by the features for classification.

\subsection{The relationship between BNCs and staged tree classifiers}

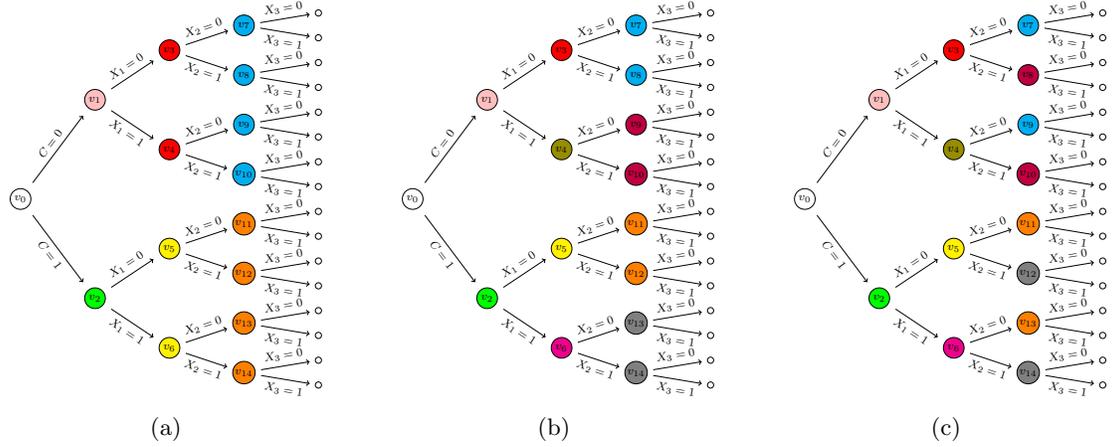
\begin{figure}
\centering
\subfloat[\label{fig:naivebnctree}]{

\centering
\scalebox{0.55}{
    \begin{tikzpicture}
\renewcommand{\xx}{1.8}
\renewcommand{\yy}{1.2}
\node (v1) at (0*\xx,0*\yy) {\stages{white}{0}};
\node (v2) at (1*\xx,2*\yy) {\stages{pink}{1}};
\node (v3) at (1*\xx,-2*\yy) {\stages{green}{2}};
\node (v4) at (2*\xx,3*\yy) {\stages{red}{3}};
\node (v5) at (2*\xx,1*\yy) {\stages{red}{4}};
\node (v6) at (2*\xx,-1*\yy) {\stages{yellow}{5}};
\node (v7) at (2*\xx,-3*\yy) {\stages{yellow}{6}};
\node (l1) at (3*\xx,3.5*\yy) {\stages{cyan}{7}};
\node (l2) at (3*\xx,2.5*\yy) {\stages{cyan}{8}};
\node (l3) at (3*\xx,1.5*\yy) {\stages{cyan}{9}};
\node (l4) at (3*\xx,0.5*\yy) {\stages{cyan}{10}};
\node (l5) at (3*\xx,-0.5*\yy) {\stages{orange}{11}};
\node (l6) at (3*\xx,-1.5*\yy) {\stages{orange}{12}};
\node (l7) at (3*\xx,-2.5*\yy) {\stages{orange}{13}};
\node (l8) at (3*\xx,-3.5*\yy) {\stages{orange}{14}};
\node (l9) at (4*\xx,3.75*\yy){\leaf};
\node (l10) at (4*\xx,3.25*\yy){\leaf};
\node (l11) at (4*\xx,2.75*\yy){\leaf};
\node (l12) at (4*\xx,2.25*\yy){\leaf};
\node (l13) at (4*\xx,1.75*\yy){\leaf};
\node (l14) at (4*\xx,1.25*\yy){\leaf};
\node (l15) at (4*\xx,0.75*\yy){\leaf};
\node (l16) at (4*\xx,0.25*\yy){\leaf};
\node (l17) at (4*\xx,-0.25*\yy){\leaf};
\node (l18) at (4*\xx,-0.75*\yy){\leaf};
\node (l19) at (4*\xx,-1.25*\yy){\leaf};
\node (l20) at (4*\xx,-1.75*\yy){\leaf};
\node (l21) at (4*\xx,-2.25*\yy){\leaf};
\node (l22) at (4*\xx,-2.75*\yy){\leaf};
\node (l23) at (4*\xx,-3.25*\yy){\leaf};
\node (l24) at (4*\xx,-3.75*\yy){\leaf};
\draw[->] (v1) --  node [above, sloped] {\scriptsize{$C=0$}} (v2);
\draw[->] (v1) -- node [below, sloped] {\scriptsize{$C=1$}}(v3);
\draw[->] (v2) --  node [above, sloped] {\scriptsize{$X_1=0$}}(v4);
\draw[->] (v2) --  node [below, sloped] {\scriptsize{$X_1=1$}}(v5);
\draw[->] (v3) --  node [above, sloped] {\scriptsize{$X_1=0$}} (v6);
\draw[->] (v3) --  node [below, sloped] {\scriptsize{$X_1=1$}} (v7);
\draw[->] (v4) --  node [above, sloped] {\scriptsize{$X_2=0$}} (l1);
\draw[->] (v4) -- node [below, sloped] {\scriptsize{$X_2=1$}}  (l2);
\draw[->] (v5) -- node [above, sloped] {\scriptsize{$X_2=0$}}  (l3);
\draw[->] (v5) -- node [below, sloped] {\scriptsize{$X_2=1$}}  (l4);
\draw[->] (v6) -- node [above, sloped] {\scriptsize{$X_2=0$}} (l5);
\draw[->] (v6) -- node [below, sloped] {\scriptsize{$X_2=1$}} (l6);
\draw[->] (v7) -- node [above, sloped] {\scriptsize{$X_2=0$}} (l7);
\draw[->] (v7) -- node [below, sloped] {\scriptsize{$X_2=1$}} (l8);
\draw[->] (l1) --  node [above, sloped] {\scriptsize{$X_3=0$}}(l9);
\draw[->] (l1) --  node [below, sloped] {\scriptsize{$X_3=1$}} (l10);
\draw[->] (l2) --  node [above, sloped] {\scriptsize{$X_3=0$}}(l11);
\draw[->] (l2) --  node [below, sloped] {\scriptsize{$X_3=1$}} (l12);
\draw[->] (l3) --  node [above, sloped] {\scriptsize{$X_3=0$}}(l13);
\draw[->] (l3) --  node [below, sloped] {\scriptsize{$X_3=1$}} (l14);
\draw[->] (l4) --  node [above, sloped] {\scriptsize{$X_3=0$}}(l15);
\draw[->] (l4) --  node [below, sloped] {\scriptsize{$X_3=1$}} (l16);
\draw[->] (l5) -- node [above, sloped] {\scriptsize{$X_3=0$}} (l17);
\draw[->] (l5) --  node [below, sloped] {\scriptsize{$X_3=1$}} (l18);
\draw[->] (l6) -- node [above, sloped] {\scriptsize{$X_3=0$}} (l19);
\draw[->] (l6) --  node [below, sloped] {\scriptsize{$X_3=1$}} (l20);
\draw[->] (l7) -- node [above, sloped] {\scriptsize{$X_3=0$}} (l21);
\draw[->] (l7) --  node [below, sloped] {\scriptsize{$X_3=1$}} (l22);
\draw[->] (l8) -- node [above, sloped] {\scriptsize{$X_3=0$}} (l23);
\draw[->] (l8) -- node [below, sloped] {\scriptsize{$X_3=1$}} (l24);
\end{tikzpicture}
}
}
\hspace{0.4cm}
\subfloat[ \label{fig:spodebnctree}]{ 
\centering
\scalebox{0.55}{
    \begin{tikzpicture}
\renewcommand{\xx}{1.8}
\renewcommand{\yy}{1.2}
\node (v1) at (0*\xx,0*\yy) {\stages{white}{0}};
\node (v2) at (1*\xx,2*\yy) {\stages{pink}{1}};
\node (v3) at (1*\xx,-2*\yy) {\stages{green}{2}};
\node (v4) at (2*\xx,3*\yy) {\stages{red}{3}};
\node (v5) at (2*\xx,1*\yy) {\stages{olive}{4}};
\node (v6) at (2*\xx,-1*\yy) {\stages{yellow}{5}};
\node (v7) at (2*\xx,-3*\yy) {\stages{magenta}{6}};
\node (l1) at (3*\xx,3.5*\yy) {\stages{cyan}{7}};
\node (l2) at (3*\xx,2.5*\yy) {\stages{cyan}{8}};
\node (l3) at (3*\xx,1.5*\yy) {\stages{purple}{9}};
\node (l4) at (3*\xx,0.5*\yy) {\stages{purple}{10}};
\node (l5) at (3*\xx,-0.5*\yy) {\stages{orange}{11}};
\node (l6) at (3*\xx,-1.5*\yy) {\stages{orange}{12}};
\node (l7) at (3*\xx,-2.5*\yy) {\stages{gray}{13}};
\node (l8) at (3*\xx,-3.5*\yy) {\stages{gray}{14}};
\node (l9) at (4*\xx,3.75*\yy){\leaf};
\node (l10) at (4*\xx,3.25*\yy){\leaf};
\node (l11) at (4*\xx,2.75*\yy){\leaf};
\node (l12) at (4*\xx,2.25*\yy){\leaf};
\node (l13) at (4*\xx,1.75*\yy){\leaf};
\node (l14) at (4*\xx,1.25*\yy){\leaf};
\node (l15) at (4*\xx,0.75*\yy){\leaf};
\node (l16) at (4*\xx,0.25*\yy){\leaf};
\node (l17) at (4*\xx,-0.25*\yy){\leaf};
\node (l18) at (4*\xx,-0.75*\yy){\leaf};
\node (l19) at (4*\xx,-1.25*\yy){\leaf};
\node (l20) at (4*\xx,-1.75*\yy){\leaf};
\node (l21) at (4*\xx,-2.25*\yy){\leaf};
\node (l22) at (4*\xx,-2.75*\yy){\leaf};
\node (l23) at (4*\xx,-3.25*\yy){\leaf};
\node (l24) at (4*\xx,-3.75*\yy){\leaf};
\draw[->] (v1) --  node [above, sloped] {\scriptsize{$C=0$}} (v2);
\draw[->] (v1) -- node [below, sloped] {\scriptsize{$C=1$}}(v3);
\draw[->] (v2) --  node [above, sloped] {\scriptsize{$X_1=0$}}(v4);
\draw[->] (v2) --  node [below, sloped] {\scriptsize{$X_1=1$}}(v5);
\draw[->] (v3) --  node [above, sloped] {\scriptsize{$X_1=0$}} (v6);
\draw[->] (v3) --  node [below, sloped] {\scriptsize{$X_1=1$}} (v7);
\draw[->] (v4) --  node [above, sloped] {\scriptsize{$X_2=0$}} (l1);
\draw[->] (v4) -- node [below, sloped] {\scriptsize{$X_2=1$}}  (l2);
\draw[->] (v5) -- node [above, sloped] {\scriptsize{$X_2=0$}}  (l3);
\draw[->] (v5) -- node [below, sloped] {\scriptsize{$X_2=1$}}  (l4);
\draw[->] (v6) -- node [above, sloped] {\scriptsize{$X_2=0$}} (l5);
\draw[->] (v6) -- node [below, sloped] {\scriptsize{$X_2=1$}} (l6);
\draw[->] (v7) -- node [above, sloped] {\scriptsize{$X_2=0$}} (l7);
\draw[->] (v7) -- node [below, sloped] {\scriptsize{$X_2=1$}} (l8);
\draw[->] (l1) --  node [above, sloped] {\scriptsize{$X_3=0$}}(l9);
\draw[->] (l1) --  node [below, sloped] {\scriptsize{$X_3=1$}} (l10);
\draw[->] (l2) --  node [above, sloped] {\scriptsize{$X_3=0$}}(l11);
\draw[->] (l2) --  node [below, sloped] {\scriptsize{$X_3=1$}} (l12);
\draw[->] (l3) --  node [above, sloped] {\scriptsize{$X_3=0$}}(l13);
\draw[->] (l3) --  node [below, sloped] {\scriptsize{$X_3=1$}} (l14);
\draw[->] (l4) --  node [above, sloped] {\scriptsize{$X_3=0$}}(l15);
\draw[->] (l4) --  node [below, sloped] {\scriptsize{$X_3=1$}} (l16);
\draw[->] (l5) -- node [above, sloped] {\scriptsize{$X_3=0$}} (l17);
\draw[->] (l5) --  node [below, sloped] {\scriptsize{$X_3=1$}} (l18);
\draw[->] (l6) -- node [above, sloped] {\scriptsize{$X_3=0$}} (l19);
\draw[->] (l6) --  node [below, sloped] {\scriptsize{$X_3=1$}} (l20);
\draw[->] (l7) -- node [above, sloped] {\scriptsize{$X_3=0$}} (l21);
\draw[->] (l7) --  node [below, sloped] {\scriptsize{$X_3=1$}} (l22);
\draw[->] (l8) -- node [above, sloped] {\scriptsize{$X_3=0$}} (l23);
\draw[->] (l8) -- node [below, sloped] {\scriptsize{$X_3=1$}} (l24);
\end{tikzpicture}
}
}
\hspace{0.4cm}
\subfloat[ \label{fig:tanbnctree}]{ 
\centering
\scalebox{0.55}{
    \begin{tikzpicture}
\renewcommand{\xx}{1.8}
\renewcommand{\yy}{1.2}
\node (v1) at (0*\xx,0*\yy) {\stages{white}{0}};
\node (v2) at (1*\xx,2*\yy) {\stages{pink}{1}};
\node (v3) at (1*\xx,-2*\yy) {\stages{green}{2}};
\node (v4) at (2*\xx,3*\yy) {\stages{red}{3}};
\node (v5) at (2*\xx,1*\yy) {\stages{olive}{4}};
\node (v6) at (2*\xx,-1*\yy) {\stages{yellow}{5}};
\node (v7) at (2*\xx,-3*\yy) {\stages{magenta}{6}};
\node (l1) at (3*\xx,3.5*\yy) {\stages{cyan}{7}};
\node (l2) at (3*\xx,2.5*\yy) {\stages{purple}{8}};
\node (l3) at (3*\xx,1.5*\yy) {\stages{cyan}{9}};
\node (l4) at (3*\xx,0.5*\yy) {\stages{purple}{10}};
\node (l5) at (3*\xx,-0.5*\yy) {\stages{orange}{11}};
\node (l6) at (3*\xx,-1.5*\yy) {\stages{gray}{12}};
\node (l7) at (3*\xx,-2.5*\yy) {\stages{orange}{13}};
\node (l8) at (3*\xx,-3.5*\yy) {\stages{gray}{14}};
\node (l9) at (4*\xx,3.75*\yy){\leaf};
\node (l10) at (4*\xx,3.25*\yy){\leaf};
\node (l11) at (4*\xx,2.75*\yy){\leaf};
\node (l12) at (4*\xx,2.25*\yy){\leaf};
\node (l13) at (4*\xx,1.75*\yy){\leaf};
\node (l14) at (4*\xx,1.25*\yy){\leaf};
\node (l15) at (4*\xx,0.75*\yy){\leaf};
\node (l16) at (4*\xx,0.25*\yy){\leaf};
\node (l17) at (4*\xx,-0.25*\yy){\leaf};
\node (l18) at (4*\xx,-0.75*\yy){\leaf};
\node (l19) at (4*\xx,-1.25*\yy){\leaf};
\node (l20) at (4*\xx,-1.75*\yy){\leaf};
\node (l21) at (4*\xx,-2.25*\yy){\leaf};
\node (l22) at (4*\xx,-2.75*\yy){\leaf};
\node (l23) at (4*\xx,-3.25*\yy){\leaf};
\node (l24) at (4*\xx,-3.75*\yy){\leaf};
\draw[->] (v1) --  node [above, sloped] {\scriptsize{$C=0$}} (v2);
\draw[->] (v1) -- node [below, sloped] {\scriptsize{$C=1$}}(v3);
\draw[->] (v2) --  node [above, sloped] {\scriptsize{$X_1=0$}}(v4);
\draw[->] (v2) --  node [below, sloped] {\scriptsize{$X_1=1$}}(v5);
\draw[->] (v3) --  node [above, sloped] {\scriptsize{$X_1=0$}} (v6);
\draw[->] (v3) --  node [below, sloped] {\scriptsize{$X_1=1$}} (v7);
\draw[->] (v4) --  node [above, sloped] {\scriptsize{$X_2=0$}} (l1);
\draw[->] (v4) -- node [below, sloped] {\scriptsize{$X_2=1$}}  (l2);
\draw[->] (v5) -- node [above, sloped] {\scriptsize{$X_2=0$}}  (l3);
\draw[->] (v5) -- node [below, sloped] {\scriptsize{$X_2=1$}}  (l4);
\draw[->] (v6) -- node [above, sloped] {\scriptsize{$X_2=0$}} (l5);
\draw[->] (v6) -- node [below, sloped] {\scriptsize{$X_2=1$}} (l6);
\draw[->] (v7) -- node [above, sloped] {\scriptsize{$X_2=0$}} (l7);
\draw[->] (v7) -- node [below, sloped] {\scriptsize{$X_2=1$}} (l8);
\draw[->] (l1) --  node [above, sloped] {\scriptsize{$X_3=0$}}(l9);
\draw[->] (l1) --  node [below, sloped] {\scriptsize{$X_3=1$}} (l10);
\draw[->] (l2) --  node [above, sloped] {\scriptsize{$X_3=0$}}(l11);
\draw[->] (l2) --  node [below, sloped] {\scriptsize{$X_3=1$}} (l12);
\draw[->] (l3) --  node [above, sloped] {\scriptsize{$X_3=0$}}(l13);
\draw[->] (l3) --  node [below, sloped] {\scriptsize{$X_3=1$}} (l14);
\draw[->] (l4) --  node [above, sloped] {\scriptsize{$X_3=0$}}(l15);
\draw[->] (l4) --  node [below, sloped] {\scriptsize{$X_3=1$}} (l16);
\draw[->] (l5) -- node [above, sloped] {\scriptsize{$X_3=0$}} (l17);
\draw[->] (l5) --  node [below, sloped] {\scriptsize{$X_3=1$}} (l18);
\draw[->] (l6) -- node [above, sloped] {\scriptsize{$X_3=0$}} (l19);
\draw[->] (l6) --  node [below, sloped] {\scriptsize{$X_3=1$}} (l20);
\draw[->] (l7) -- node [above, sloped] {\scriptsize{$X_3=0$}} (l21);
\draw[->] (l7) --  node [below, sloped] {\scriptsize{$X_3=1$}} (l22);
\draw[->] (l8) -- node [above, sloped] {\scriptsize{$X_3=0$}} (l23);
\draw[->] (l8) -- node [below, sloped] {\scriptsize{$X_3=1$}} (l24);
\end{tikzpicture}
}
}
\caption{Representation of BNCs as staged trees classifiers: (a): naive BNC in Figure \ref{fig:naivebn} as a staged tree classifier; (b): SPODE BNC in Figure \ref{fig:spodebn} as a staged tree classifier; (c): TAN BNC in Figure \ref{fig:tanbn} as a staged tree classifier. All variables are assumed binary. \label{fig:bnctree}}
\end{figure}

The BNCs reviewed in Section \ref{sec:bnc} can now be represented as staged tree classifiers. Since a BNC is a BN with a DAG $G$, one can construct its equivalent staged tree $T_G$  as in Section \ref{sec:noia}.  For instance naive BNCs (Figure \ref{fig:naivebnctree}), 
SPODE BNCs (Figure \ref{fig:spodebnctree}) and TAN BNCs (Figure \ref{fig:tanbnctree}) 
can concisely be represented as staged tree classifiers. However, the class of staged trees classifiers is much larger than that of BNCs, as formalized in Proposition~\ref{prop:1}. For a class $C$ and features $\bm{X}$, let $\mathcal{M}_{\mathcal{G}}^{\textnormal{C}}$ be the space of BNCs (where $C$ is the root), and $\mathcal{M}_T^{\textnormal{C}}$ the space of $(C,\bm{X})$-compatible staged trees.




\begin{proposition}
\label{prop:1}
$\mathcal{M}_{\mathcal{G}}^{\textnormal{C}}\subset \mathcal{M}_T^{\textnormal{C}}$.
\end{proposition}

The proof follows from \citet{Varando2021}.

In particular naive Bayes, SPODE, and TAN classifiers are all 
staged tree classifiers with a specific staging structure as 
described next for the naive Bayes.  Recall that the set $V_k$ includes the nodes of the tree at distance $k$ from the root and let $T(v)$ be the subtree of $T$ rooted at a vertex $v$.



\begin{proposition}
\label{prop:2}
        Let $G$ be the DAG of a naive BNC. Then $T_G$ is a $(C,\bm{X})$-compatible staged tree where, for all $v \in V_1$, the subtree $T_G(v)$ 
	is a $\bm{X}$-compatible staged tree where all nodes at the same distance from the root are in the 
		same stage. 
		In particular, $T_G$
		has 
		 $|\mathcal{C}|$ stages per each feature. 

\end{proposition}


\subsection{Conditional independence in staged tree classifiers}

For the specific task of classification, it is possible to derive two results about the dependence between the features and the class in staged tree classifiers.

\begin{proposition}
\label{prop:3}
	If all $v\in V_k$ of a staged tree classifier are in the same stage then 
	$(C, X_1, X_2, \ldots, X_{k-1})$ and $X_k$ are marginally independent, i.e. 
	$(C, X_1, \ldots, X_{k-1})\independent X_k$.
\end{proposition}

\begin{proposition}
\label{prop:4}
	If for all $v \in V_1$,  $T(v)$ has the same stage 
	structure over the vertices at distance $k-2$ from the root, 
	then $X_k$ is independent of $C$ conditionally on 
	$X_1,\dots,X_{k-1}$, i.e. $C\independent X_k\,|\, X_1,\dots,X_{k-1}$.
\end{proposition}

\begin{figure}
\centering
\subfloat[ \label{fig:ci1}]{ 
\centering
\scalebox{0.7}{
    \begin{tikzpicture}
\renewcommand{\xx}{1.8}
\renewcommand{\yy}{1.2}
\node (v1) at (0*\xx,0*\yy) {\stages{white}{0}};
\node (v2) at (1*\xx,2*\yy) {\stages{pink}{1}};
\node (v3) at (1*\xx,-2*\yy) {\stages{green}{2}};
\node (v4) at (2*\xx,3*\yy) {\stages{red}{3}};
\node (v5) at (2*\xx,1*\yy) {\stages{olive}{4}};
\node (v6) at (2*\xx,-1*\yy) {\stages{yellow}{5}};
\node (v7) at (2*\xx,-3*\yy) {\stages{magenta}{6}};
\node (l1) at (3*\xx,3.5*\yy) {\stages{cyan}{7}};
\node (l2) at (3*\xx,2.5*\yy) {\stages{cyan}{8}};
\node (l3) at (3*\xx,1.5*\yy) {\stages{cyan}{9}};
\node (l4) at (3*\xx,0.5*\yy) {\stages{cyan}{10}};
\node (l5) at (3*\xx,-0.5*\yy) {\stages{cyan}{11}};
\node (l6) at (3*\xx,-1.5*\yy) {\stages{cyan}{12}};
\node (l7) at (3*\xx,-2.5*\yy) {\stages{cyan}{13}};
\node (l8) at (3*\xx,-3.5*\yy) {\stages{cyan}{14}};
\node (l9) at (4*\xx,3.75*\yy){\leaf};
\node (l10) at (4*\xx,3.25*\yy){\leaf};
\node (l11) at (4*\xx,2.75*\yy){\leaf};
\node (l12) at (4*\xx,2.25*\yy){\leaf};
\node (l13) at (4*\xx,1.75*\yy){\leaf};
\node (l14) at (4*\xx,1.25*\yy){\leaf};
\node (l15) at (4*\xx,0.75*\yy){\leaf};
\node (l16) at (4*\xx,0.25*\yy){\leaf};
\node (l17) at (4*\xx,-0.25*\yy){\leaf};
\node (l18) at (4*\xx,-0.75*\yy){\leaf};
\node (l19) at (4*\xx,-1.25*\yy){\leaf};
\node (l20) at (4*\xx,-1.75*\yy){\leaf};
\node (l21) at (4*\xx,-2.25*\yy){\leaf};
\node (l22) at (4*\xx,-2.75*\yy){\leaf};
\node (l23) at (4*\xx,-3.25*\yy){\leaf};
\node (l24) at (4*\xx,-3.75*\yy){\leaf};
\draw[->] (v1) --  node [above, sloped] {\scriptsize{$C=0$}} (v2);
\draw[->] (v1) -- node [below, sloped] {\scriptsize{$C=1$}}(v3);
\draw[->] (v2) --  node [above, sloped] {\scriptsize{$X_1=0$}}(v4);
\draw[->] (v2) --  node [below, sloped] {\scriptsize{$X_1=1$}}(v5);
\draw[->] (v3) --  node [above, sloped] {\scriptsize{$X_1=0$}} (v6);
\draw[->] (v3) --  node [below, sloped] {\scriptsize{$X_1=1$}} (v7);
\draw[->] (v4) --  node [above, sloped] {\scriptsize{$X_2=0$}} (l1);
\draw[->] (v4) -- node [below, sloped] {\scriptsize{$X_2=1$}}  (l2);
\draw[->] (v5) -- node [above, sloped] {\scriptsize{$X_2=0$}}  (l3);
\draw[->] (v5) -- node [below, sloped] {\scriptsize{$X_2=1$}}  (l4);
\draw[->] (v6) -- node [above, sloped] {\scriptsize{$X_2=0$}} (l5);
\draw[->] (v6) -- node [below, sloped] {\scriptsize{$X_2=1$}} (l6);
\draw[->] (v7) -- node [above, sloped] {\scriptsize{$X_2=0$}} (l7);
\draw[->] (v7) -- node [below, sloped] {\scriptsize{$X_2=1$}} (l8);
\draw[->] (l1) --  node [above, sloped] {\scriptsize{$X_3=0$}}(l9);
\draw[->] (l1) --  node [below, sloped] {\scriptsize{$X_3=1$}} (l10);
\draw[->] (l2) --  node [above, sloped] {\scriptsize{$X_3=0$}}(l11);
\draw[->] (l2) --  node [below, sloped] {\scriptsize{$X_3=1$}} (l12);
\draw[->] (l3) --  node [above, sloped] {\scriptsize{$X_3=0$}}(l13);
\draw[->] (l3) --  node [below, sloped] {\scriptsize{$X_3=1$}} (l14);
\draw[->] (l4) --  node [above, sloped] {\scriptsize{$X_3=0$}}(l15);
\draw[->] (l4) --  node [below, sloped] {\scriptsize{$X_3=1$}} (l16);
\draw[->] (l5) -- node [above, sloped] {\scriptsize{$X_3=0$}} (l17);
\draw[->] (l5) --  node [below, sloped] {\scriptsize{$X_3=1$}} (l18);
\draw[->] (l6) -- node [above, sloped] {\scriptsize{$X_3=0$}} (l19);
\draw[->] (l6) --  node [below, sloped] {\scriptsize{$X_3=1$}} (l20);
\draw[->] (l7) -- node [above, sloped] {\scriptsize{$X_3=0$}} (l21);
\draw[->] (l7) --  node [below, sloped] {\scriptsize{$X_3=1$}} (l22);
\draw[->] (l8) -- node [above, sloped] {\scriptsize{$X_3=0$}} (l23);
\draw[->] (l8) -- node [below, sloped] {\scriptsize{$X_3=1$}} (l24);
\end{tikzpicture}
}

}
\hspace{2cm}
\subfloat[ \label{fig:ci2}]{ 
\centering
\scalebox{0.7}{
    \begin{tikzpicture}
\renewcommand{\xx}{1.8}
\renewcommand{\yy}{1.2}
\node (v1) at (0*\xx,0*\yy) {\stages{white}{0}};
\node (v2) at (1*\xx,2*\yy) {\stages{pink}{1}};
\node (v3) at (1*\xx,-2*\yy) {\stages{green}{2}};
\node (v4) at (2*\xx,3*\yy) {\stages{red}{3}};
\node (v5) at (2*\xx,1*\yy) {\stages{olive}{4}};
\node (v6) at (2*\xx,-1*\yy) {\stages{yellow}{5}};
\node (v7) at (2*\xx,-3*\yy) {\stages{magenta}{6}};
\node (l1) at (3*\xx,3.5*\yy) {\stages{cyan}{7}};
\node (l2) at (3*\xx,2.5*\yy) {\stages{purple}{8}};
\node (l3) at (3*\xx,1.5*\yy) {\stages{orange}{9}};
\node (l4) at (3*\xx,0.5*\yy) {\stages{gray}{10}};
\node (l5) at (3*\xx,-0.5*\yy) {\stages{cyan}{11}};
\node (l6) at (3*\xx,-1.5*\yy) {\stages{purple}{12}};
\node (l7) at (3*\xx,-2.5*\yy) {\stages{orange}{13}};
\node (l8) at (3*\xx,-3.5*\yy) {\stages{gray}{14}};
\node (l9) at (4*\xx,3.75*\yy){\leaf};
\node (l10) at (4*\xx,3.25*\yy){\leaf};
\node (l11) at (4*\xx,2.75*\yy){\leaf};
\node (l12) at (4*\xx,2.25*\yy){\leaf};
\node (l13) at (4*\xx,1.75*\yy){\leaf};
\node (l14) at (4*\xx,1.25*\yy){\leaf};
\node (l15) at (4*\xx,0.75*\yy){\leaf};
\node (l16) at (4*\xx,0.25*\yy){\leaf};
\node (l17) at (4*\xx,-0.25*\yy){\leaf};
\node (l18) at (4*\xx,-0.75*\yy){\leaf};
\node (l19) at (4*\xx,-1.25*\yy){\leaf};
\node (l20) at (4*\xx,-1.75*\yy){\leaf};
\node (l21) at (4*\xx,-2.25*\yy){\leaf};
\node (l22) at (4*\xx,-2.75*\yy){\leaf};
\node (l23) at (4*\xx,-3.25*\yy){\leaf};
\node (l24) at (4*\xx,-3.75*\yy){\leaf};
\draw[->] (v1) --  node [above, sloped] {\scriptsize{$C=0$}} (v2);
\draw[->] (v1) -- node [below, sloped] {\scriptsize{$C=1$}}(v3);
\draw[->] (v2) --  node [above, sloped] {\scriptsize{$X_1=0$}}(v4);
\draw[->] (v2) --  node [below, sloped] {\scriptsize{$X_1=1$}}(v5);
\draw[->] (v3) --  node [above, sloped] {\scriptsize{$X_1=0$}} (v6);
\draw[->] (v3) --  node [below, sloped] {\scriptsize{$X_1=1$}} (v7);
\draw[->] (v4) --  node [above, sloped] {\scriptsize{$X_2=0$}} (l1);
\draw[->] (v4) -- node [below, sloped] {\scriptsize{$X_2=1$}}  (l2);
\draw[->] (v5) -- node [above, sloped] {\scriptsize{$X_2=0$}}  (l3);
\draw[->] (v5) -- node [below, sloped] {\scriptsize{$X_2=1$}}  (l4);
\draw[->] (v6) -- node [above, sloped] {\scriptsize{$X_2=0$}} (l5);
\draw[->] (v6) -- node [below, sloped] {\scriptsize{$X_2=1$}} (l6);
\draw[->] (v7) -- node [above, sloped] {\scriptsize{$X_2=0$}} (l7);
\draw[->] (v7) -- node [below, sloped] {\scriptsize{$X_2=1$}} (l8);
\draw[->] (l1) --  node [above, sloped] {\scriptsize{$X_3=0$}}(l9);
\draw[->] (l1) --  node [below, sloped] {\scriptsize{$X_3=1$}} (l10);
\draw[->] (l2) --  node [above, sloped] {\scriptsize{$X_3=0$}}(l11);
\draw[->] (l2) --  node [below, sloped] {\scriptsize{$X_3=1$}} (l12);
\draw[->] (l3) --  node [above, sloped] {\scriptsize{$X_3=0$}}(l13);
\draw[->] (l3) --  node [below, sloped] {\scriptsize{$X_3=1$}} (l14);
\draw[->] (l4) --  node [above, sloped] {\scriptsize{$X_3=0$}}(l15);
\draw[->] (l4) --  node [below, sloped] {\scriptsize{$X_3=1$}} (l16);
\draw[->] (l5) -- node [above, sloped] {\scriptsize{$X_3=0$}} (l17);
\draw[->] (l5) --  node [below, sloped] {\scriptsize{$X_3=1$}} (l18);
\draw[->] (l6) -- node [above, sloped] {\scriptsize{$X_3=0$}} (l19);
\draw[->] (l6) --  node [below, sloped] {\scriptsize{$X_3=1$}} (l20);
\draw[->] (l7) -- node [above, sloped] {\scriptsize{$X_3=0$}} (l21);
\draw[->] (l7) --  node [below, sloped] {\scriptsize{$X_3=1$}} (l22);
\draw[->] (l8) -- node [above, sloped] {\scriptsize{$X_3=0$}} (l23);
\draw[->] (l8) -- node [below, sloped] {\scriptsize{$X_3=1$}} (l24);
\end{tikzpicture}
}

}
\caption{Staged trees classifiers embedding conditional independence statements between features and the class. (a): $X_3\independent C$; (b): $X_3 \independent C \, | \, X_1, X_2$.\label{fig:ci}}
\end{figure}
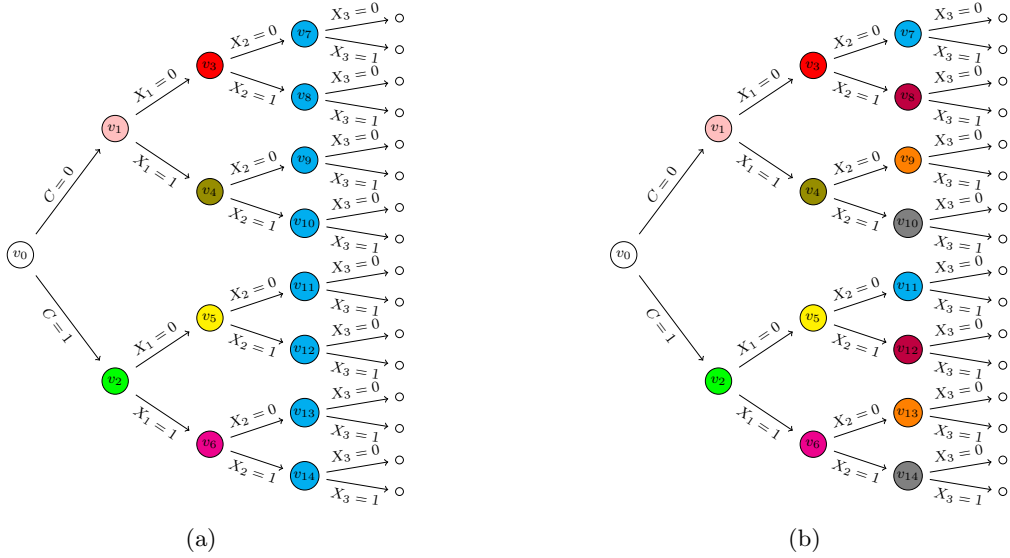

These two results are illustrated in Figure \ref{fig:ci}. For instance, consider the features associated to the last random variable in Figure \ref{fig:ci2}. The vertices in the upper half are framed as the vertices in the bottom half thus implying that the class variable is conditionally independent of the last feature given all others.



\subsection{Naive staged tree classifiers}
\label{sec:naive}
The class of staged tree classifiers is extremely rich and for any 
classification task the number of candidate models that could explain the 
relationship between class and features increases exponentially. 
One first common assumption that we make here is to consider only 
$(C,\bm{X})$-compatible staged trees, ones where only vertices at the same distance 
from the root can be in a same stage. 
However, even with this assumption the model class of staged tree classifiers is 
still much richer than that of BNCs. Therefore, just like BNCs whose DAGs have 
restricted topologies (as SPODE and TAN classifiers) have been studied, 
next we introduce a class of simpler staged tree classifiers.

We have discussed that in many practical applications the naive BNC has very good 
classification performance despite of its simplicity. 
A naive BNC has a total of $\sum_{j=1}^n|\mathbb{C}|\cdot(|\mathbb{X}_j|-1)+|\mathbb{C}|-1$ free parameters that need to be learnt, whilst its DAG is always fixed. 
Similarly, we  introduce a class of staged tree classifiers which has the constraint of having the same number of free parameters as naive BNCs, and therefore has the same complexity, whilst being a much richer class of models then the naive BNC. 

\begin{definition}
A $(C,\bm{X})$-compatible staged tree classifier such that for every $k\leq p$, the set 
	$V_k$ is partitioned into $|\mathbb{C}|$ stages is called \emph{naive}.  
\end{definition}

It staightforwardly follows from the definition that naive staged tree classifiers have the same number of free parameters as naive Bayes classifiers.

 Despite of the strict constraint on the number of parameters, the class of naive staged tree classifiers is still rich and extends naive BNCs in a 
non-trivial way. More formally, let $\mathcal{M}_{\mathcal{G}}^{\textnormal{naive}}$ and  $\mathcal{M}_T^{\textnormal{naive}}$  be the space of naive BNCs and of naive staged tree classifiers, respectively, for a class $C$ and features $\bm{X}$.

\begin{proposition}
$\mathcal{M}_{\mathcal{G}}^{\textnormal{naive}}\subset \mathcal{M}_{T}^{\textnormal{naive}}$.
\end{proposition}

 Differently to naive BNCs, it is not sufficient to simply learn the probabilities 
of the naive staged tree classifier, 
but also the staging structure has to be discovered.   However, because of the strict restriction on the number of parameters, fast algorithms can be devised to efficiently explore the model space. Notice that in a binary classification problem the set $V_k$ must be partitioned into two subsets. 





Critically and differently to naive Bayes classifiers, naive staged trees are capable of representing complex decision rules.  For instance, consider the simplest scenario of a binary class with two binary features. The naive staged tree classifier in Figure \ref{fig:staged2} (assuming the first variable is the class variable), which, as already noticed, does not have a naive BNC representation,  is capturing the only 2-XOR present.

To investigate further the capabilities of naive staged tree classifiers in expressing complex decision rules, we simulate $N_{train} = 200$ observations from $n = 10$ binary variables 
$X_1, \ldots, X_{10}$ 
taking values in $\mathbb{X} = \{-1,+1\}^n$. 
We define the class variable as the parity (or XOR) function 
$C = \prod_{i=1}^n X_i$ and compare naive Bayes, 
random forest and naive staged tree classifiers over $N_{test} = 10000$ test 
instances, obtaining the results in Table~\ref{tab:xorex}. 
See Section \ref{sec:naive} for details on the learning of naive staged tree classifiers.
As expected, the Naive Bayes 
classifier is unable to represent the parity function \citep{Varando2015} and wrongly classifies
the class in more than $40\%$ of the test data. 
Similar performances are obtained by random forests (implemented with the \texttt{randomForest} R package using $500$ trees), even if theoretically 
they have much larger expressive power. 
Conversely, the naive staged tree correctly learns the parity function with an accuracy of 90\%.

\begin{table}
	\center
	\begin{tabular}{ccccccccc}
		\toprule
		&  \multicolumn{2}{c}{ST\_Naive} & & \multicolumn{2}{c}{RFor}& &\multicolumn{2}{c}{NB}  \\ 
		\midrule
		& -1 & 1 & & -1 & 1 & & -1 & 1 \\
		-1 & 0.4623 & 0.0713 & &  0.3972 & 0.3123 & & 0.4314 & 0.3982 \\
		1 &0.0347 & 0.4317 & & 0.0998 & 0.1907 & & 0.0656 & 0.1048 \\
		\bottomrule
\end{tabular}
\caption{Proportion of predicted instances in the simulated XOR example for the naive staged tree classifier (ST\_Naive), random forest (RFor) and naive Bayes (NB).} 
	\label{tab:xorex}
\end{table}

\section{Learning staged tree classifiers}
\label{sec:learning}
The learning of the structure of a staged tree from data is challenging due to the exponential increase of the size of the tree with the number of random variables. The first learning algorithm used for this purpose was the agglomerative hierarchical clustering proposed by \citet{Freeman2011}. Other learning algorithms were then introduced by \citet{Silander2013} and \citet{Collazo2016}, among others.
Recently, a staged trees implementation 
in \texttt{R} was made available in the \texttt{stagedtrees} package~\citep{Carli2020}
with various searching algorithms to estimate stage structures from data.

In this article we present different methods for learning staged 
tree classifiers. 
All methods follow three main steps: 
(i) an optimal order of the variables is identified; 
(ii) the full probability tree based on all random variables is pruned to 
speed-up computations; 
(iii) model search heuristics are used to identify high-scoring structures. 
We next give details about each of these phases. Furthermore, in Section \ref{sec:naivealg} we introduce algorithms specifically designed for naive staged tree classifiers.

\subsection{Choosing a features' order}
\label{sec:order}

It is well known that the ordering of the variables affect the quality of a statistical classifier \citep[e.g.][]{Hruschka2007}.  
Thus, choosing the order of the features in a staged tree classifier is expected to 
be important in order to obtain good performances.  

\begin{figure}
\begin{center}
	\includegraphics{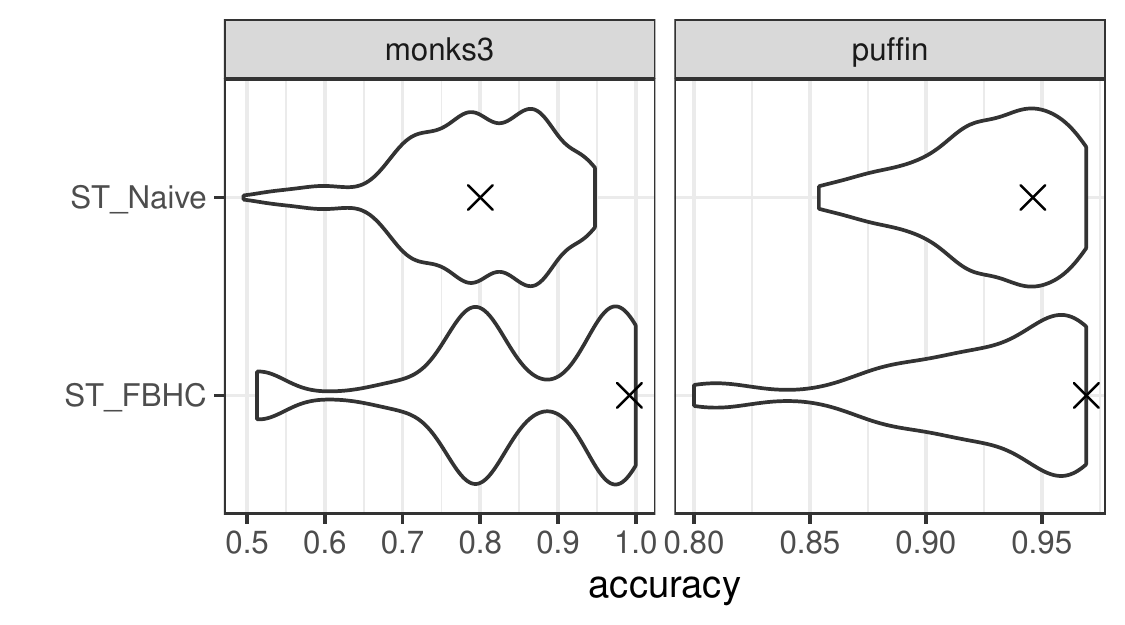}
	\caption{Distribution of accuracy for different orders of 
	         features. 
                 Accuracies obtained with the CMI ordering are shown with a cross. 
		 Results using two learning algorithms for datasets \texttt{puffin} 
		 and \texttt{monks3}.} 
		 \label{fig:order}
\end{center}
\end{figure}

To confirm this, we perform an empirical study where we compute
classification accuracies for different staged tree classifiers over all possible
features' orders. Given the combinatorial explosion of the number of orders 
we limit this study to two illustrative datasets with a small amount of features, namely 
\texttt{puffin} and \texttt{monks3} (see Table \ref{table:data} for details). 
From  Figure~\ref{fig:order} we can observe that, as expected, the order of
the features is highly relevant with respect to classification performance. 
For instance, the naive staged tree learnt with hierarchical clustering, ST\_Naive
(as described in Section~\ref{sec:naivealg}), and the staged tree classifier learnt with the fast backward-hill climbing 
algorithm, ST\_FBHC (maximizing BIC score, see Section~\ref{sec:learn}), exhibit a variety of accuracies (from $0.5$ to $1$) for the same dataset. 

It is therefore critical to couple any algorithm to learn a staged tree
with an appropriate method to select a good ordering of the variables. 
For staged tree classifiers, we tested various ordering heuristics and we 
propose here the use of the conditional mutual information (CMI) criterion.
The order of the features given by the CMI criterion is obtained by iteratively 
selecting the feature that maximise the conditional mutual information with respect to 
the class variable, given the features previously selected.
The performance of such ordering is shown with a cross symbol in Figure~\ref{fig:order}. We can observe
that in both \texttt{puffin} and \texttt{monks3} problems, the CMI order leads to
a staged tree which perfrom better than the majority of the possible orders.

\subsection{Dealing with unobserved instances}
\label{sec:unobserved}   
Once the order of the variables is selected, it is possible that some of the nodes in the tree
are not observed in the training data. 
For such nodes, estimating the associated probabilities is not possible without 
further assumptions. 
As implemented in the \texttt{R} package \texttt{stagedtrees}, for each feature, unobserved
situations are joined in a common stage, called \textit{unobserved stage}, and a uniform probability is 
imposed for their distribution. Furthermore, such unobserved stages are 
excluded from the stage structure search. 
This has the effect of drastically reducing the number of nodes for which a
stage structure must be learnt, thus critically reducing training time. 

From a statistical point of view, this does not affect the number of free parameters of the corresponding statistical model, since probability distributions of these unobserved stages have not to be estimated. This procedure can be seen as the usual pruning step of learning algorithms for classification trees. From the naive staged tree classifier perspective, it follows that the number of stages in the tree is unaffected, since its complexity, i.e. number of free parameters, does not change.

\subsection{Learning algorithms}
\label{sec:learn} 

Different heuristic algorithms can be used to search the space of
possible stage structures. 
We rely on the available implementations in the \texttt{stagedtrees} package, in 
particular we use greedy approaches that maximise the BIC score \citep{Gorgen2020} or 
iterative methods that join nodes into the same stages based on 
the distance between their associated probabilities.  
More details of the algorithms are described in \cite{Carli2020} and 
in the \texttt{stagedtrees} documentation. 

\subsection{Learning naive staged tree classifiers}
\label{sec:naivealg}

As described in Section~\ref{sec:naive}, a naive staged tree classifier is a 
staged tree where the class is the first variable in the tree and 
nodes at the same distance from the root are assigned to a fixed number of stages equal to
the number of possible values of the class (e.g. two for a binary classification 
problem). 

The two algorithms based on clustering (hierarchical and k-means) available
in the \texttt{stagedtrees} package can be used to learn naive staged tree classifiers. They cluster probabilities into a user-selected number of stages which, for the purposes of classification, can be fixed to number of possible values of the class variable.

\section{Experimental study}
\label{sec:experimental}

\begin{table}
\begin{center}
\begin{tabular}{cccc}
\toprule
Dataset & \# observations & \# variables & \# atomic events \\
\midrule
\texttt{asym} &100&4&16\\
\texttt{breastCancer} &683&10&1024\\
\texttt{chestSim500} &500&8&256\\
\texttt{energy1} &768&9&1728\\
\texttt{energy2} &768&9&1728\\
\texttt{fallEld} &5000&4&64\\
\texttt{fertility} &100&10&15552\\
\texttt{monks1} &432&7&864\\
\texttt{monks2}&432&7&864\\
\texttt{monks3} &432&7&864\\
\texttt{puffin} &69&6&768\\
\texttt{ticTacToe} &958&10&39366\\
\texttt{titanic} &2201&4&32\\
\texttt{voting} &435&17&131072\\
\bottomrule
\end{tabular}
\end{center}
\caption{Details about the 14 datasets included in the experimental study. \label{table:data}}
\end{table}
The classification accuracy for binary classification of staged tree classifiers is investigated in a comprehensive simulation study involving 14 datasets, whose details are given in Table \ref{table:data}. Each dataset is randomly divided ten times in train set ($80\%$ of the data) to learn the classifiers and test set (remaining $20\%$) to predict the response. The reported performance measures, area under the curve (AUC) and balanced accuracy, are computed as the mean over the ten replications. 

First, 9 model search algorithms to learn staged tree classifiers are compared, namely: ST\_BHC (backward hill-climbing); ST\_BJ\_01 (backward joining of vertices that have Kullback-Leibler divergence between their floret probability distributions less than 0.01); ST\_BJ\_20 (as ST\_BJ\_01 but with threshold at 0.20); ST\_FBHC (a fast backward hill-climbing where two vertices are joined whenever the score is increased); ST\_Full (each vertex is in its own stage); ST\_HC\_Full (hill-climbing algorithm starting from ST\_Full); ST\_HC\_Indep (hill-climbing algorithm starting from a tree where all vertices associated to the same variable are in the same stage); ST\_Naive\_HC (naive staged tree learnt with hierarchical clustering); ST\_Naive\_KM (naive staged tree learnt with k-means). Further details about these algorithms can be found in \citet{Carli2020}. Due to computational restrictions, for ST\_HC\_Full the model search is restricted to the first five features according to the variable ordering chosen through CMI, whilst for ST\_BHC and ST\_HC\_Indep only the first seven are considered. The vertices corresponding to the remaining variables are still used for classification but left as in the starting tree of the model search.

\begin{figure}
\begin{center}
\includegraphics[scale=0.6]{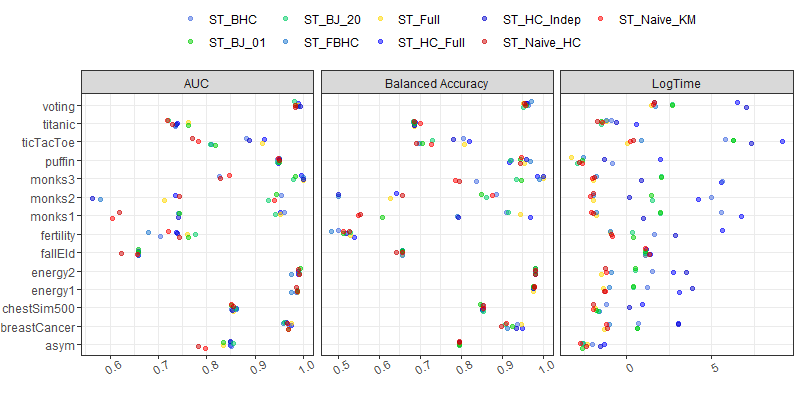}
\end{center}
	\caption{AUC, balanced accuracy and logarithm of time spent 
	 for structure learning for nine staged tree classifiers algorithms over fourteen datasets.\label{fig:treesc}}
\end{figure}

The results of the experiment are reported in Figure \ref{fig:treesc}, which suggests the following conclusions:
\begin{itemize}
\item The ST\_Full model (in yellow), which does not require any model search and has the largest number of parameters, has in general lower AUC and balanced accuracy than other staged trees. This highlights the need of a model-based search of simpler models;
\item Models based on hill-climbing (in blue) overall perform better than others (in particular ST\_HC\_Full). This is expected since these are the most refined learning algorithms and, as a consequence, they are also the slowest. 
\item Models based on backward joining (in green) have a satisfactory performance, often comparable to that of hill-climbing models, whilst being much quicker to learn.
\item Naive staged trees (in red) can be learnt extremely quickly and whilst often they have a lower performance, there are cases where they are comparable to the one of much more complex trees (see e.g. the balanced accuracy for the \texttt{titanic} dataset)
\end{itemize}

\begin{figure}
\begin{center}
\includegraphics[scale=0.6]{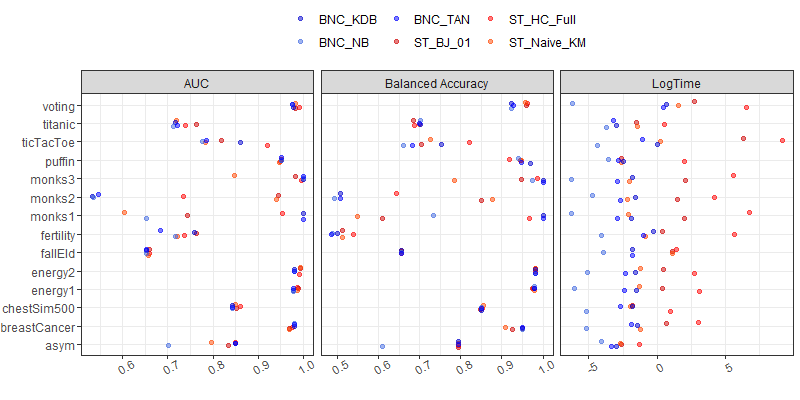}
\end{center}
\caption{AUC, balanced accuracy and logarithm of time spent for structure learning for three staged tree classifiers (in red) and three BNCs (in blue) over fourteen datasets. \label{fig:bnsc}}
\end{figure}

Next, we compare staged trees classifiers with their competitor generative classifier, namely BNCs. For ease of exposition, three representative staged trees are selected (ST\_BJ\_01, ST\_HC\_Full and ST\_Naive\_KM) and three BNCs are fitted: (i) the TAN BNC (BNC\_TAN); (ii) the 3-dependence BNC (BNC\_KDB); (iii) the naive Bayes (BNC\_NB). The results are reported in Figure \ref{fig:bnsc}. We can see that for most datasets there is one staged tree classifier (in red) that outperforms BNCs (in blue). Due to the complexity of the models, staged trees are in general slower to learn, but the ST\_Naive\_KM, due to its simplicity, has learning times comparable to those of generic BNCs.

\begin{figure}
\begin{center}
\includegraphics[scale=0.6]{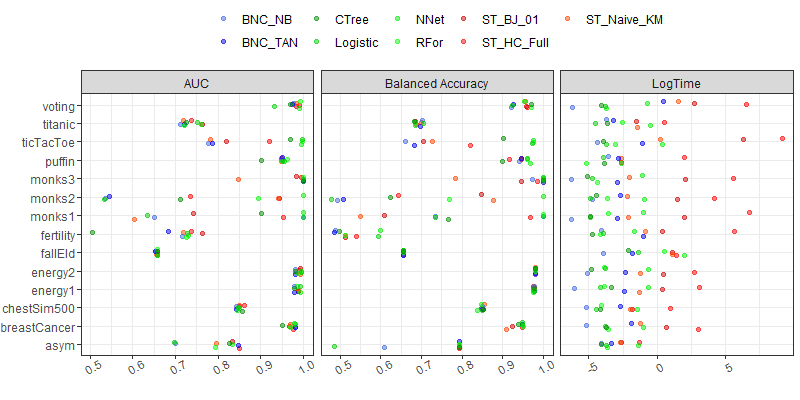}
\end{center}
\caption{AUC, balanced accuracy and logarithm of time spent for structure learning for three staged tree classifiers (in red), two BNCs (in blue) and other discriminative models (in green) over fourteen datasets.\label{fig:allc}}
\end{figure}

Figure \ref{fig:allc} reports the results of the simulation experiments for three staged tree classifiers (ST\_BJ\_01, ST\_HC\_Full and ST\_Naive\_KM) as well as other state-of-the-art generative and discriminative classifiers, namely: (i) the naive Bayes (BNC\_NB); (ii) the TAN BNC (BNC\_TAN); (iii) Classification trees (CTree) (iv) Logistic regression (Logistic); (v) Neural Networks with 20 hidden layers and 0.01 weight decay (NNet); (vi) Random Forests combining 100 classification trees (RFor). Although in some cases discriminative classifiers (in green) outperform staged trees (in red), in many others they have comparable AUC and balanced accuracy. However, as shown in the next section, staged tree classifiers have the capability of producing an understanding of the relationship between the class and the features, since they are generative models. As already noticed, staged trees have an advantage over BNCs (in blue). Although the learning time for generic staged trees is larger, the learning time for naive staged tree classifiers is comparable to that of state-of-the-art classifiers.

\begin{figure}
\begin{center}
\includegraphics[scale=0.6]{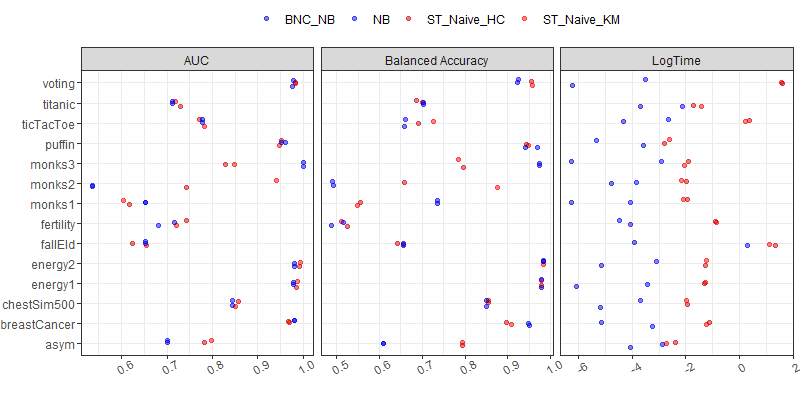}
\end{center}
\caption{AUC, balanced accuracy and logarithm of time spent for structure learning for two naive staged tree classifiers (in red) and two implementations of naive Bayes classifiers (in blue) over fourteen datasets.\label{fig:naivec}}
\end{figure}

Last, we compare the performance of naive Bayes classifiers with the one of naive staged tree classifiers in Figure \ref{fig:naivec}. The overall conclusion is that in most cases naive staged trees (in red) outperform naive Bayes in terms of AUC and balanced accuracy. Furthermore, although naive staged trees require more learning time since their structure has to be discovered, these can be learnt very quickly and most times in less than one second.

\section{An example of a staged tree classifier}
\label{sec:applied}
To illustrate the capabilities of staged trees classifiers we next develop an example classification analysis over the freely available \texttt{titanic} dataset, which provides information on the fate of the Titanic passengers. It has three binary variables (Survived, Sex and Age) and a categorical variable Class taking four levels (1st, 2nd, 3rd and Crew). The aim is to correctly classify whether the Titanic passengers survived or not based on their gender, age and travelling class.

\begin{figure}
\begin{center}
\includegraphics[scale=0.6]{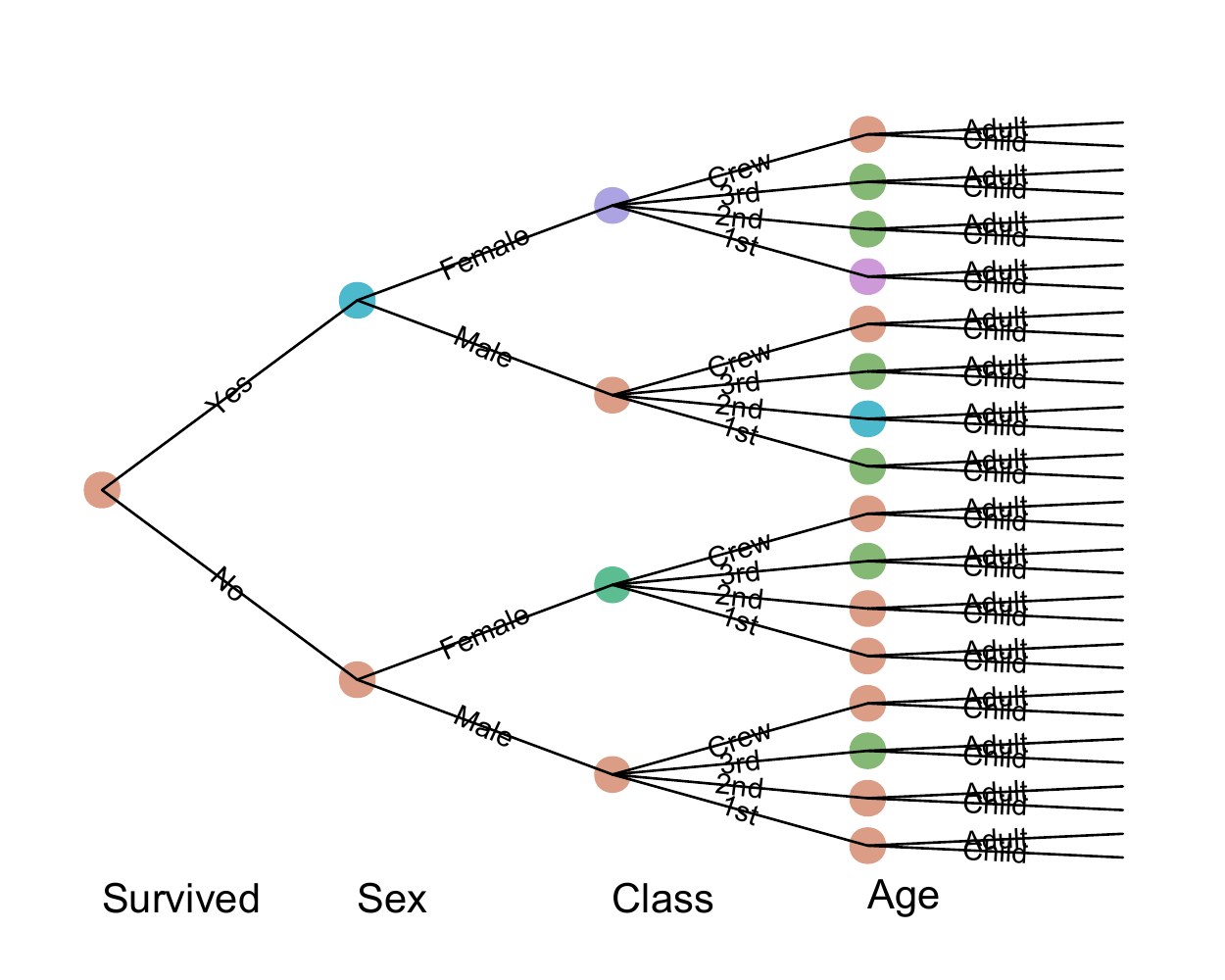}
\end{center}
\caption{Staged tree classifier ST\_BJ\_01 learnt over the full Titanic dataset. \label{fig:tit1}}
\end{figure}

From Figure \ref{fig:treesc} we can see that one of the best staged tree classifiers is the ST\_BJ\_01 learnt using a backward joining of the vertices based on the Kullback-Leibler divergence and a threshold of 0.1. In Figure \ref{fig:tit1} we report the staged tree classifier ST\_BJ\_01 learnt over the full Titanic dataset using the \texttt{R} package \texttt{stagedtrees}. By investigating the staging structure we can deduce conditional independence statements relating to the classification variable (Survived) and the features. From stages associated to Class we can deduce that $P(\textnormal{Class}\,|\, \textnormal{Sex} = \textnormal{Male}, \textnormal{Survived})= P(\textnormal{Class}\,|\,\textnormal{Sex} = \textnormal{Male})$ since the second and the fourth vertices (starting from the top) are in the same stage. This implies the asymmetric conditional independence
\[
\textnormal{Class} \independent \textnormal{Survived} \,|\, \textnormal{Sex} = \textnormal{Male}
\]
The complex staging structure over the Age variable also implies asymmetric conditional independences. We can notice that all paths going through an edge labelled Crew are in the same stage for the variable Age. This implies that 
\[
\textnormal{Age} \independent \textnormal{Survived} \,|\, \textnormal{Class} = \textnormal{Crew}
\]
The same conclusion can also be drawn for Class $=$ 3rd.

\begin{figure}
\begin{center}
\includegraphics[scale = 0.6]{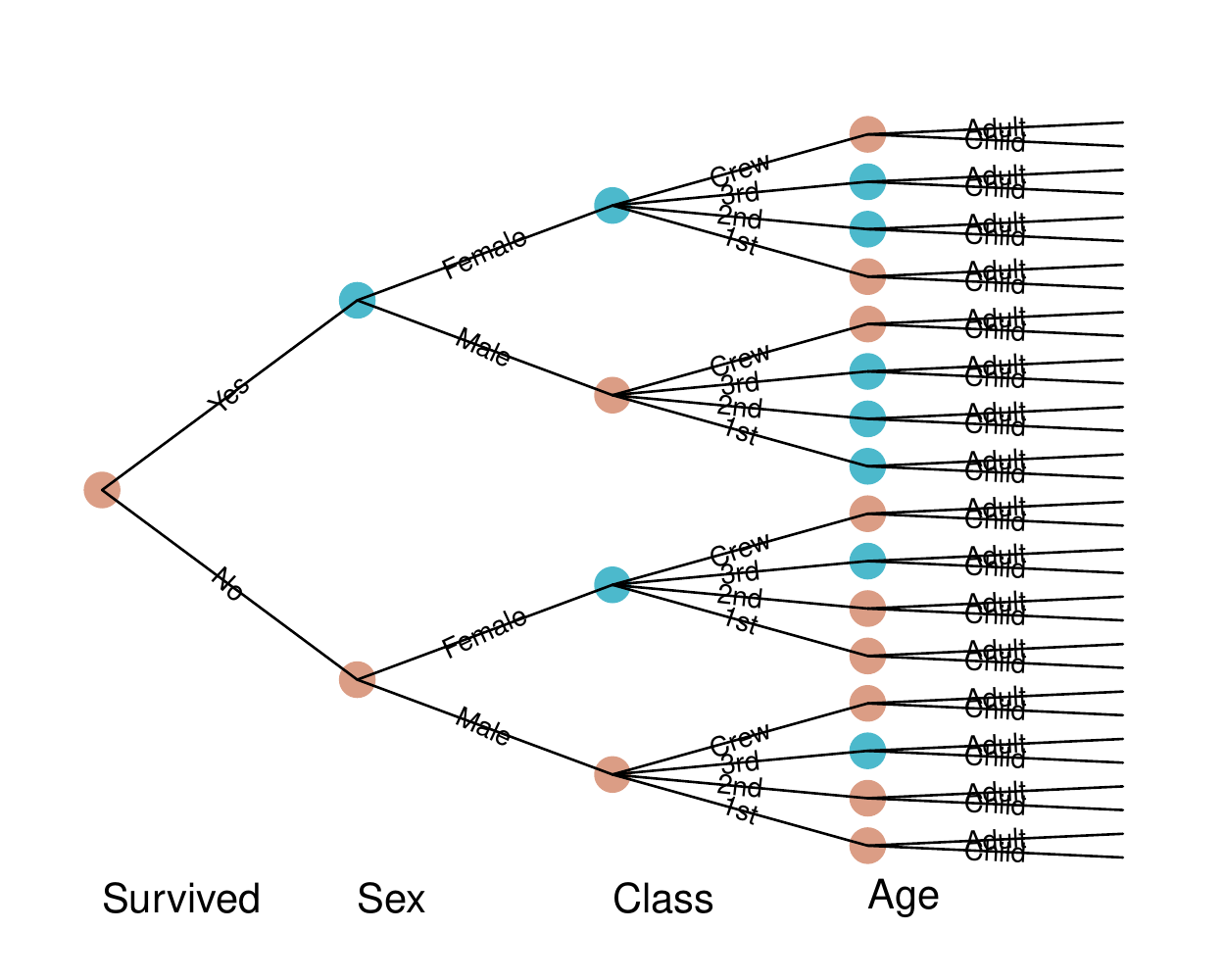}
\end{center}
\caption{Naive staged tree classifier learnt over the full Titanic dataset using the \texttt{stages\_kmeans} algorithm. \label{fig:tit2}}
\end{figure}

As an additional illustration in Figure \ref{fig:tit2} is reported the naive staged tree classifier learnt over the full Titanic dataset using the k-means hierarchical clustering algorithm. The staging structure over the variables Sex and Class implies that Sex and Survived are not independent and that Class is conditionally independent of Survived given Sex. 
The staging structure over the Age variable is a lot more complex describing highly asymmetric constraints on the associated probabilities. Whilst imposing much more flexible dependence structures, the naive staged tree classifier has the same complexity of the naive Bayes classifier, meaning they have the same number of independent parameters.

BNCs of different complexity are also learnt over the full \texttt{titanic} dataset using the \texttt{bnclassify} \texttt{R} package. Irrespective of the complexity chosen, the model selection search always returns the simple naive Bayes classifier. Given that staged tree classifiers outperfom BNCs in classification measures for the Titanic dataset (see Figure \ref{fig:bnsc}), as well as for other datasets, this observations highlights the need of context-specific generative classifiers that can more flexibly model the dependence structure between the classification variable and the features.

\section{Discussion}
\label{sec:conclusions}
Staged trees classifiers are a highly-expressive new class of generative classifiers with classification performance comparable to that of state-of-the-art classifiers. They embed context-specific conditional independence statements which can be easily read by the staging of the vertices of the tree. These are implemented in the freely available \texttt{stagedtrees} \texttt{R} package.

A special staged tree classifier is the naive staged tree classifier which, whilst having the same complexity as the naive Bayes classifier, can flexibly represent complex classification rule. Naive staged trees not only relax the assumption of conditional indepedence of the features as in naive Bayes classifiers but also have better performances in classification, as highlighted by the simulation study.

Naive staged trees are learnt from data using a clustering algorithm of the probability distributions over the non-leaf vertices of the tree. Such algorithms divide the vertices at the same distance from the root in $|\mathbb{C}|$ stages. More generally, we could devise clustering algorithms were, for each variable, the number of stages is automatically selected according to an optimality criterion. The development of these algorithms is the focus of ongoing research.

Furthermore, many model search algorithms implemented in \texttt{stagedtrees} are based on the maximization of a model score, by the default the negative BIC. We are currently investigating algorithms based on the minimization of the classification error, as commonly implemented for BNCs. 

\section*{Acknowledgments} 
Gherardo Varando's work was partly funded by the European Research
Council (ERC) Synergy Grant “Understanding and Modelling the Earth
System with Machine Learning (USMILE)” under Grant Agreement No 855187.



\bibliography{Bib}

\begin{thebibliography}{45}
\providecommand{\natexlab}[1]{#1}
\providecommand{\url}[1]{\texttt{#1}}
\expandafter\ifx\csname urlstyle\endcsname\relax
  \providecommand{\doi}[1]{doi: #1}\else
  \providecommand{\doi}{doi: \begingroup \urlstyle{rm}\Url}\fi

\bibitem[Barclay et~al.(2013)Barclay, Hutton, and Smith]{Barclay2013}
L.~Barclay, J.~Hutton, and J.~Smith.
\newblock Refining a {B}ayesian network using a chain event graph.
\newblock \emph{International Journal of Approximate Reasoning}, 54:\penalty0
  1300--1309, 2013.

\bibitem[Benjumeda et~al.(2019)Benjumeda, Luengo-Sanchez, Larranaga, and
  Bielza]{Benjumeda2019}
M.~Benjumeda, S.~Luengo-Sanchez, P.~Larranaga, and C.~Bielza.
\newblock {Tractable learning of Bayesian networks from partially observed
  data}.
\newblock \emph{Pattern Recognition}, 91:\penalty0 190--199, 2019.

\bibitem[Bielza and Larra\~{n}aga(2014)]{Bielza2014}
A.~Bielza and P.~Larra\~{n}aga.
\newblock Discrete {B}ayesian network classifiers: a survey.
\newblock \emph{ACM Computing Surveys}, 47\penalty0 (1):\penalty0 5:1--5:43,
  2014.

\bibitem[Boutilier et~al.(1996)Boutilier, Friedman, Goldszmidt, and
  Koller]{Boutilier1996}
C.~Boutilier, N.~Friedman, M.~Goldszmidt, and D.~Koller.
\newblock Context-specific independence in {Bayesian} networks.
\newblock In \emph{Proceedings of the 12th Conference on Uncertainty in
  Artificial Intelligence}, pages 115--123, 1996.

\bibitem[Breiman et~al.(1984)Breiman, Friedman, Olshen, and Stone]{Breiman1984}
L.~Breiman, J.~Friedman, R.~Olshen, and C.~Stone.
\newblock \emph{Classification and regression trees}.
\newblock Wadsworth, 1984.

\bibitem[Cano et~al.(2012)Cano, G\'{o}mez-Olmedo, Moral, P\'{e}rez-Ariza, and
  Salmer\'{o}n]{Cano2012}
A.~Cano, M.~G\'{o}mez-Olmedo, S.~Moral, C.~P\'{e}rez-Ariza, and
  A.~Salmer\'{o}n.
\newblock Learning recursive probability trees from probabilistic potentials.
\newblock \emph{International Journal of Approximate Reasoning}, 53\penalty0
  (9):\penalty0 1367--1387, 2012.

\bibitem[Carli et~al.(2020)Carli, Leonelli, Riccomagno, and Varando]{Carli2020}
F.~Carli, M.~Leonelli, E.~Riccomagno, and G.~Varando.
\newblock {The R package stagedtrees for structural learning of stratified
  staged trees}.
\newblock \emph{Journal of Statistical Software}, 102\penalty0 (6):\penalty0
  1--30, 2020.

\bibitem[Collazo and Smith(2016)]{Collazo2016}
R.~Collazo and J.~Smith.
\newblock A new family of non-local priors for chain event graph model
  selection.
\newblock \emph{Bayesian Analysis}, 11\penalty0 (4):\penalty0 1165--1201, 2016.

\bibitem[Collazo et~al.(2018)Collazo, G\"{o}rgen, and Smith]{Collazo2018}
R.~Collazo, C.~G\"{o}rgen, and J.~Smith.
\newblock \emph{Chain event graphs}.
\newblock Chapmann \& Hall, 2018.

\bibitem[Domingos and Pazzani(1997)]{Domingos1997}
P.~Domingos and M.~Pazzani.
\newblock On the optimality of the simple {B}ayesian network classifiers.
\newblock \emph{Machine Learning}, 29\penalty0 (2-3):\penalty0 103--130, 1997.

\bibitem[Flores et~al.(2012)Flores, G{\'a}mez, and Mart{\'\i}nez]{Flores2012}
M.~Flores, J.~G{\'a}mez, and A.~Mart{\'\i}nez.
\newblock Supervised classification with {B}ayesian networks: a review on
  models and applications.
\newblock In \emph{Intelligent data analysis for real-life applications: theory
  and practice}, pages 72--102. 2012.

\bibitem[Freeman and Smith(2011)]{Freeman2011}
G.~Freeman and J.~Smith.
\newblock Bayesian {MAP} model selection of chain event graphs.
\newblock \emph{Journal of Multivariate Analysis}, 102\penalty0 (7):\penalty0
  1152--1165, 2011.

\bibitem[Friedman et~al.(1997)Friedman, Geiger, and Goldszmidt]{Friedman1997}
N.~Friedman, D.~Geiger, and M.~Goldszmidt.
\newblock Bayesian network classifiers.
\newblock \emph{Machine Learning}, 29\penalty0 (2-3):\penalty0 131--163, 1997.

\bibitem[Geiger and Heckerman(1996)]{Geiger1996}
D.~Geiger and D.~Heckerman.
\newblock Knowledge representation and inference in similarity networks and
  {B}ayesian multinets.
\newblock \emph{Artificial Intelligence}, 82:\penalty0 45--74, 1996.

\bibitem[Genewein et~al.(2020)Genewein, McGrath, Delétang, Mikulik, Martic,
  Legg, and Ortega]{deep}
T.~Genewein, T.~McGrath, G.~Delétang, V.~Mikulik, M.~Martic, S.~Legg, and
  P.~Ortega.
\newblock Algorithms for causal reasoning in probability trees.
\newblock \emph{arXiv:2010.12237}, 2020.

\bibitem[G{\"o}rgen et~al.(2015)G{\"o}rgen, Leonelli, and Smith]{Gorgen2015}
C.~G{\"o}rgen, M.~Leonelli, and J.~Smith.
\newblock A differential approach for staged trees.
\newblock In \emph{European Conference on Symbolic and Quantitative Approaches
  to Reasoning and Uncertainty}, pages 346--355. Springer, 2015.

\bibitem[Gurwicz and Lerner(2006)]{Gurwicz2006}
Y.~Gurwicz and B.~Lerner.
\newblock Bayesian class-matched multinet classifier.
\newblock In \emph{Proceedings of the Joint IAPR International Conference on
  Structural, Syntactic, and Statistical Pattern Recognition}, pages 145--153,
  2006.

\bibitem[Görgen et~al.(2020)Görgen, Leonelli, and Marigliano]{Gorgen2020}
C.~Görgen, M.~Leonelli, and O.~Marigliano.
\newblock The curved exponential family of a staged tree.
\newblock \emph{Electronic Journal of Statistics}, 16\penalty0 (1):\penalty0
  2607--2620, 2020.

\bibitem[Ho(1995)]{Ho1995}
T.~Ho.
\newblock Random decision forests.
\newblock In \emph{Proceedings of the 3rd International Conference on Document
  Analysis and Recognition}, volume~1, pages 278--282, 1995.

\bibitem[Hruschka~Jr and Ebecken(2007)]{Hruschka2007}
E.~R. Hruschka~Jr and N.~F. Ebecken.
\newblock {Towards efficient variables ordering for Bayesian networks
  classifier}.
\newblock \emph{Data \& Knowledge Engineering}, 63\penalty0 (2):\penalty0
  258--269, 2007.

\bibitem[Huang et~al.(2003)Huang, King, and Lyu]{Huang2003}
K.~Huang, I.~King, and M.~Lyu.
\newblock Discriminative training of {Bayesian Chow-Liu} multinet classifiers.
\newblock In \emph{Proceedings of the International Joint Conference on Neural
  Networks}, pages 484--488, 2003.

\bibitem[Hussein and Santos(2004)]{Hussein2004}
A.~Hussein and E.~Santos.
\newblock Exploring case-based {B}ayesian networks and {B}ayesian multi-nets
  for classification.
\newblock In \emph{Proceedings of the 17th Conference of the Canadian Society
  of Computational Studies of Intelligence}, pages 485--492, 2004.

\bibitem[Jaeger et~al.(2006)Jaeger, Nielsen, and Silander]{Jaeger2006}
M.~Jaeger, J.~Nielsen, and T.~Silander.
\newblock Learning probabilistic decision graphs.
\newblock \emph{International Journal of Approximate Reasoning}, 42\penalty0
  (1-2):\penalty0 84--100, 2006.

\bibitem[Keeble et~al.(2017)Keeble, Thwaites, Baxter, Barber, Parslow, and
  Law]{Keeble2017}
C.~Keeble, P.~Thwaites, P.~Baxter, S.~Barber, R.~Parslow, and G.~Law.
\newblock Learning through chain event graphs: The role of maternal factors in
  childhood type 1 diabetes.
\newblock \emph{American Journal of Epidemiology}, 186\penalty0 (10):\penalty0
  1204--1208, 2017.

\bibitem[Keogh and Pazzani(2002)]{Keogh2002}
E.~Keogh and M.~Pazzani.
\newblock Learning the structure of augmented {B}ayesian classifiers.
\newblock \emph{International Journal on Artificial Intelligence Tools},
  11\penalty0 (4):\penalty0 587--601, 2002.

\bibitem[Leonelli(2019)]{Leonelli2019}
M.~Leonelli.
\newblock Sensitivity analysis beyond linearity.
\newblock \emph{International Journal of Approximate Reasoning}, 113:\penalty0
  106--118, 2019.

\bibitem[Leonelli and Varando(2021)]{Leonelli2021}
M.~Leonelli and G.~Varando.
\newblock Context-specific causal discovery for categorical data using staged
  trees.
\newblock \emph{arXiv:2106.04416}, 2021.

\bibitem[Leonelli and Varando(2022{\natexlab{a}})]{Leonelli2022a}
M.~Leonelli and G.~Varando.
\newblock Highly efficient structural learning of sparse staged trees.
\newblock \emph{arXiv:2206.06970}, 2022{\natexlab{a}}.

\bibitem[Leonelli and Varando(2022{\natexlab{b}})]{Leonelli2022b}
M.~Leonelli and G.~Varando.
\newblock Structural learning of simple staged trees.
\newblock \emph{arXiv:2203.04390}, 2022{\natexlab{b}}.

\bibitem[Leonelli et~al.(2017)Leonelli, G{\"o}rgen, and Smith]{Leonelli2017}
M.~Leonelli, C.~G{\"o}rgen, and J.~Smith.
\newblock Sensitivity analysis in multilinear probabilistic models.
\newblock \emph{Information Sciences}, 411:\penalty0 84--97, 2017.

\bibitem[Ling and Zhang(2002)]{Ling2002}
C.~X. Ling and H.~Zhang.
\newblock The representational power of discrete {B}ayesian networks.
\newblock \emph{Journal of Machine Learning Research}, 3:\penalty0 709--721,
  2002.

\bibitem[Mihaljevic et~al.(2018)Mihaljevic, Bielza, and
  Larra\~{n}aga]{Mihaljevic2018}
B.~Mihaljevic, C.~Bielza, and P.~Larra\~{n}aga.
\newblock \emph{bnclassify: learning discrete {B}ayesian network classifiers
  from data}, 2018.
\newblock R package version 0.4.0.

\bibitem[Minsky(1961)]{Minsky1961}
M.~Minsky.
\newblock Steps towards artificial intelligence.
\newblock In \emph{Computers and Thought}, pages 406--450, 1961.

\bibitem[O'Donnell(2014)]{Donnell2014}
R.~O'Donnell.
\newblock \emph{Analysis of {B}oolean functions}.
\newblock Cambridge University Press, 2014.

\bibitem[Pensar et~al.(2015)Pensar, Nyman, Koski, and Corander]{Pensar2015}
J.~Pensar, H.~Nyman, T.~Koski, and J.~Corander.
\newblock Labeled directed acyclic graphs: a generalization of context-specific
  independence in directed graphical models.
\newblock \emph{Data Mining and Knowledge Discovery}, 29\penalty0 (2):\penalty0
  503--533, 2015.

\bibitem[Pensar et~al.(2016)Pensar, Nyman, Lintusaari, and
  Corander]{Pensar2016}
J.~Pensar, H.~Nyman, J.~Lintusaari, and J.~Corander.
\newblock The role of local partial independence in learning of {B}ayesian
  networks.
\newblock \emph{International Journal of Approximate Reasoning}, 69:\penalty0
  91--105, 2016.

\bibitem[Poole and Zhang(2003)]{Poole2003}
D.~Poole and N.~Zhang.
\newblock Exploiting contextual independence in probabilistic inference.
\newblock \emph{Journal of Artificial Intelligence Research}, 18:\penalty0
  263--313, 2003.

\bibitem[Shafer(1996)]{Shafer1996}
G.~Shafer.
\newblock \emph{The art of causal conjecture}.
\newblock MIT Press, 1996.

\bibitem[Silander and Leong(2013)]{Silander2013}
T.~Silander and T.-Y. Leong.
\newblock A dynamic programming algorithm for learning chain event graphs.
\newblock In \emph{Proceedings of the International Conference on Discovery
  Science}, pages 201--216, 2013.

\bibitem[Smith and Anderson(2008)]{Smith2008}
J.~Smith and P.~Anderson.
\newblock Conditional independence and chain event graphs.
\newblock \emph{Artificial Intelligence}, 172\penalty0 (1):\penalty0 42 -- 68,
  2008.

\bibitem[Specht(1990)]{Specht1990}
D.~Specht.
\newblock Probabilistic neural networks.
\newblock \emph{Neural Networks}, 3\penalty0 (1):\penalty0 109--118, 1990.

\bibitem[Thwaites et~al.(2010)Thwaites, Smith, and Riccomagno]{Thwaites2010}
P.~Thwaites, J.~Smith, and E.~Riccomagno.
\newblock Causal analysis with chain event graphs.
\newblock \emph{Artificial Intelligence}, 174\penalty0 (12-13):\penalty0
  889--909, 2010.

\bibitem[Varando et~al.(2015)Varando, Bielza, and Larra\~{n}aga]{Varando2015}
G.~Varando, C.~Bielza, and P.~Larra\~{n}aga.
\newblock Decision boundary for discrete {B}ayesian network classifiers.
\newblock \emph{Journal of Machine Learning Research}, 16:\penalty0 2725--2749,
  2015.

\bibitem[Varando et~al.(2016)Varando, Bielza, and Larra{\~n}aga]{Varando2016}
G.~Varando, C.~Bielza, and P.~Larra{\~n}aga.
\newblock {Decision functions for chain classifiers based on Bayesian networks
  for multi-label classification}.
\newblock \emph{International Journal of Approximate Reasoning}, 68:\penalty0
  164--178, 2016.

\bibitem[Varando et~al.(2021)Varando, Carli, and Leonelli]{Varando2021}
G.~Varando, F.~Carli, and M.~Leonelli.
\newblock Staged trees and asymmetry-labeled dags.
\newblock \emph{arXiv:2108.01994}, 2021.

\end{thebibliography}





\end{document}